\documentclass[preprint,12pt]{elsarticle}

\usepackage{hyperref}
\usepackage{float}
\usepackage{verbatim}
\usepackage{multirow}
\usepackage{xcolor}

\restylefloat{figure}
\restylefloat{table}

\usepackage{array}
\usepackage{booktabs}
\usepackage{adjustbox}
\usepackage{ltablex}
\usepackage{graphicx, caption, subcaption}

\usepackage{amsfonts}
\usepackage{enumitem}

\usepackage{amssymb}
\usepackage{amsmath}

\begin{document}

\begin{frontmatter}



\title{Three-dimensional Deep Shape Optimization with a Limited Dataset}


\author[1,2]{Yongmin Kwon}

\affiliation[1]{organization={Cho Chun Shik Graduate School of Mobility, Korea Advanced Institute of Science and Technology},
            addressline={193, Munji-ro}, 
            city={Yuseong-gu},
            postcode={34051}, 
            state={Daejeon},
            country={Republic of Korea}}
            
\author[1,2]{Namwoo Kang\corref{cor1}}
\affiliation[2]{organization={Narnia Labs},
            addressline={193, Munji-ro}, 
            city={Yuseong-gu},
            postcode={34051}, 
            state={Daejeon},
            country={Republic of Korea}}
\cortext[cor1]{Corresponding author}

\begin{abstract}
{Generative models have attracted considerable attention for their ability to produce novel shapes. However, their application in mechanical design remains constrained due to the limited size and variability of available datasets. This study proposes a deep learning-based optimization framework specifically tailored for shape optimization with limited datasets, leveraging positional encoding and a Lipschitz regularization term to robustly learn geometric characteristics and maintain a meaningful latent space. Through extensive experiments, the proposed approach demonstrates robustness, generalizability and effectiveness in addressing typical limitations of conventional optimization frameworks. The validity of the methodology is confirmed through multi-objective shape optimization experiments conducted on diverse three-dimensional datasets, including wheels and cars, highlighting the model’s versatility in producing practical and high-quality design outcomes even under data-constrained conditions.}
\end{abstract}

\begin{graphicalabstract}
\includegraphics[width=1.0\textwidth]{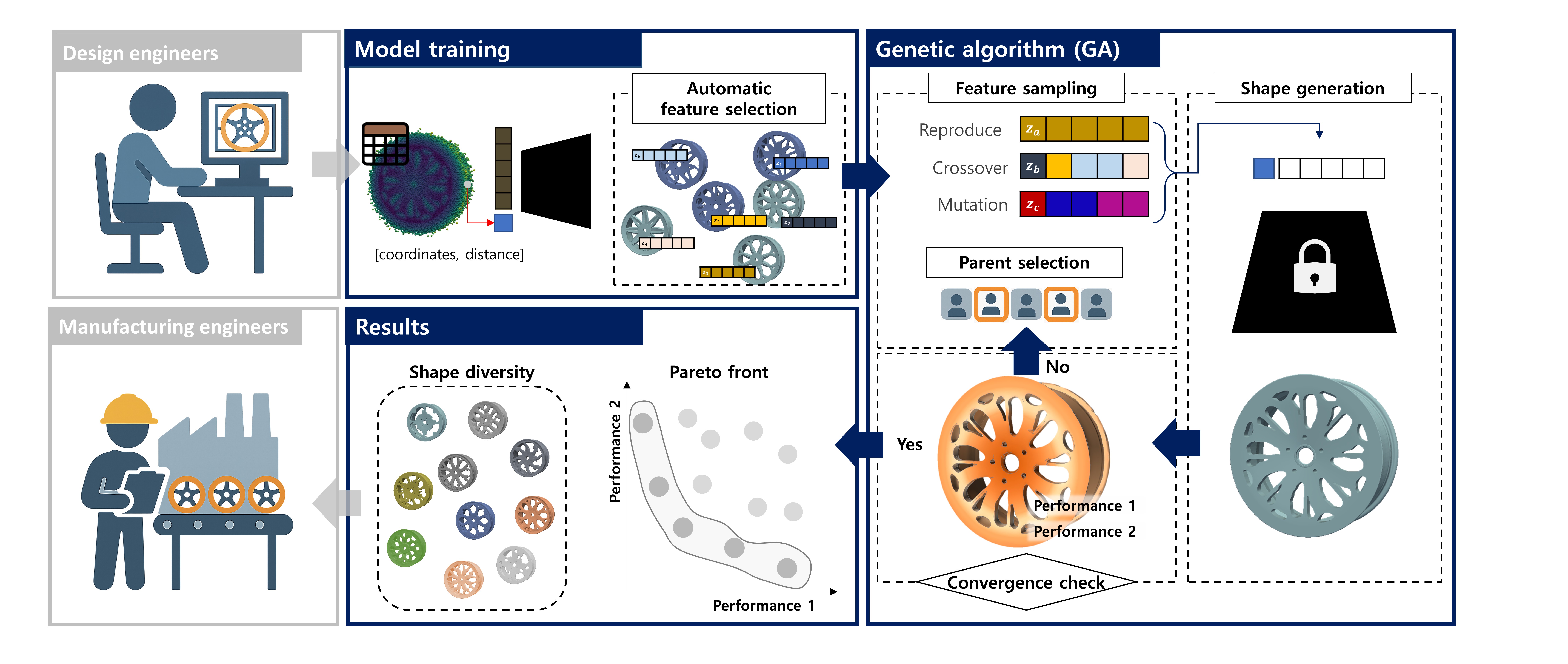}
\end{graphicalabstract}

\begin{highlights}

\item Overcomes explicit parameterization limits using data-driven shape extraction.
\item Optimizes 3D shapes effectively with a limited dataset approach.
\item Achieves diverse design outcomes in multi-objective optimization.
\item Validates the advanced performance of the framework through multiple experiments.
\end{highlights}


\end{frontmatter}

\section{Introduction}
\label{introduction}

Artificial intelligence (AI) has shown significant achievements in various domains, including regression, classification and segmentation \unskip~\cite{regenwetter2022deep, xu2025two}. Building upon these advancements, research in mechanical design, such as computer-aided design (CAD) and computer-aided engineering, has increasingly utilized AI to overcome the limitations of traditional methodologies and provide new insights \unskip~\cite{li2022machine, munoz2020aerodynamic, belhadj2022optimization, li2019surrogate, kim2022deep, akbar2024simulation, akbar2024heat, alghamdi2024double, sher2024computational}. Deep generative models (DGM), such as the variational autoencoder (VAE), the generative adversarial network (GAN) and the diffusion model, have garnered increasing attention due to their ability to learn from data and create new data that were not previously accessible \unskip~\cite{goodfellow2014generative, kingma2013auto, ho2020denoising}. This characteristic is a significant solution to the persistent data scarcity problem in the mechanical domain. Unlike conventional generative design approaches, DGM guarantees data diversity, facilitating the generation of design suggestions that incorporate engineering functionality \unskip~\cite{regenwetter2022deep}.

Traditional parametric shape optimization produces different optimal results depending on how feature selection is performed. This difference becomes more pronounced for complex shapes with highly dependent parameters. In contrast, AI models can automatically analyze the features of data through learning and compress data into a latent space. From an optimal design perspective, the latent space can be used as design variables, providing automatic feature selection functionality \unskip~\cite{deb2002fast, cunningham2020sparsity}.

One of the significant issues in employing DGM for mechanical design lies in the need for varied data within the same category. Mechanical parts, which are fundamentally mass-produced, do not have abundant data within the same design domain and existing data show only minimal variation \unskip~\cite{hong2025deepjeb, li2025generative}. These intrinsic data characteristics create significant challenges for AI when attempting to identify subtle distinctions within a single category, as it tends to focus on the overall form, which can lead to issues like mode collapse.

This study aims to propose a three-dimensional (3D) deep learning model tailored for optimal design applications. The proposed model automatically selects features from generative AI training data while preserving shape diversity. Furthermore, the proposed model demonstrated robust performance on limited datasets, addressing a bottleneck previously identified in optimal design approaches. This research introduces a versatile shape optimization framework using AI models \unskip~\cite{oh2019deep, lee2023t}. This study investigates shape optimization across distinct datasets, an endeavor that conventional parametric techniques and existing DGM-based approaches have yet to achieve. In addition, this framework is applied to shape optimization problems, demonstrating its universality through optimizations performed via computational fluid dynamics (CFD) and finite element method (FEM) simulations.

This paper is structured as follows: Section~\ref{sec:rel_works} introduces previous research on shape optimization, topology optimization and shape optimization using DGM and discusses data representation and deep learning architecture in the context of 3D data. Section~\ref{sec:method} provides an overview of the entire framework of this study, from data collection to shape optimization. Section~\ref{sec:exp} examines the originality and validity of the methodology through three shape optimization experiments. Finally, Section~\ref{sec:conclusion} presents conclusions and directions for future work.

\section{Related Works}
\label{sec:rel_works}

\subsection{Shape Optimization Method}
\label{method}

\subsubsection{Parametric Shape Optimization}

Data generated through CAD programs consists of combinations of various geometric variables, which can be utilized as design parameters in the optimization process. Parametric shape optimization is a design method in which design parameters, within a predefined range, satisfy constraints based on a defined CAD model while simultaneously identifying the optimal shape to minimize the objective function \unskip~\cite{harries2018upfront,agarwal2019enhancing,robinson2012optimizing}. This method is widely used in various engineering fields, including automotive, aerospace, electromagnetic and acoustic \unskip~\cite{li2022machine, munoz2020aerodynamic,lu2023acoustic,chen2024reduced}.

Parametric shape optimization has advanced with surrogate modeling for predicting engineering performance because it allows for defining data within a specified range via design parameters. In particular, substantial progress has been made in areas where the data can be easily defined by simple curves or encapsulated by numerical characteristics \unskip~\cite{li2022machine, munoz2020aerodynamic, belhadj2022optimization, li2019surrogate, kim2022deep, akbar2024simulation,akbar2024heat}. Experimental results indicate that the extensive design space and intricate parameter dependencies enable heuristic optimization methods, such as the Non-dominated Sorting Genetic Algorithm II (NSGA-2), to outperform gradient-based approaches \unskip~\cite{miguel2012shape, bagazinski2023ship}.

For non-parametric data such as 3D meshes, parameterization can be performed using techniques like polycube mapping or free-form deformation \unskip~\cite{umetani2018learning, salmoiraghi2018free, gomez2024analysis, gagnon2012two}. These methods can be implemented relatively easily and transform complex and detailed shapes. Furthermore, when constructing surrogate models, mesh data can be represented as a graph and processed with graph neural networks, enabling the solution of more generalized problems \unskip~\cite{baque2018geodesic}. However, a limitation of these methods is that, depending on the resolution of the defined parameters, capturing fine details becomes challenging and the number of parameters increases.

\subsubsection{Topology Optimization-based Generative Design}

Topology optimization is the most effective method for generating shapes that satisfy prescribed boundary conditions and meet specific objective functions \unskip~\cite{bendsoe1989optimal, bendsoe1999material, sigmund2013topology, liu2014efficient, deng2021efficient}. This methodology has undergone numerous advancements since introducing the initial solid isotropic material with penalization approach in the academic literature.

It is the most widely utilized method in the field of generative design because it can generate new data based on constraints defined by designers \unskip~\cite{mcknight2017generative, na2021study, jang2022generative, shin2023topology}. In generative design, topology optimization explores new shapes throughout the design process, from problem definition to detailed design. This approach streamlines the process by numerically evaluating and generating optimal designs, producing better outcomes than traditional stage-by-stage methods.

Generative models that leverage topology optimization excel in data generation under multiple boundary conditions, allowing the production of various designs. However, a notable drawback is that they tend to yield many similar data instances \unskip~\cite{matejka2018dream}. To overcome these limitations, recent studies have increasingly focused on combining topology optimization with deep generative models. For example, Oh et al. used topology optimization to create seed data to train a deep generative model \unskip~\cite{oh2019deep}. They then trained a GAN model to generate diverse wheel designs with various spoke configurations. Furthermore, there are emerging cases that incorporate reinforcement learning to address design diversity and multi-objective requirements \unskip~\cite{jang2022generative,yang2024multi}.

\subsubsection{Deep Generative Model-based Shape Optimization}
\label{sec:dgm-rel}

Recently, research on deep generative models has been advancing significantly. Extensive work is being carried out on creative generative models such as GANs, VAEs and diffusion models, which are capable of generating new images and shapes \unskip~\cite{goodfellow2014generative, kingma2013auto, ho2020denoising}. Deep generative models have evolved according to the type of data involved. For instance, they are particularly effective for processing non-parametric data such as images or 3D meshes and extracting low-dimensional features. In the field of optimal design, research is being conducted on generating and optimizing shapes using deep generative models.

Optimization studies utilizing DGMs have seen significant advancements in domains that are well-suited for image representation, such as meta structures or airfoils, which are relatively easy to parameterize \unskip~\cite{kou2023aeroacoustic, chen2020airfoil, selig1996uiuc, on2024novel}. In particular, new architectures have been developed to better define airfoils, typically represented by Bézier curves, to improve the feasibility of the generated data \unskip~\cite{chen2020airfoil}.

For 3D data, architectures based on implicit neural representations, as introduced by Park et al., have been widely adopted \unskip~\cite{park2019deepsdf}. These models are highly versatile, serving not only as generative models, but also as surrogate models. Their ability to perform auto-differentiation makes sensitivity analysis straightforward, making them well-suited for capturing local variations. This approach has been particularly useful for vehicle aerodynamic optimization problems, which typically require high computational resources \unskip~\cite{park2019deepsdf, remelli2020meshsdf, rosset2023interactive}.

Overall, data-driven deep generative models distinguish themselves from topology optimization-based generative design in that they can explore more realistic data distributions and discover unique configurations. However, they require large amounts of data.

\begin{table}[tbh]
    {
    \caption{{Comparison with other optimization methods}}
    \label{tab:comparison}
    \addtocounter{table}{-1}
    \begin{adjustbox}{minipage=1.4\textwidth, center}
        \begin{tabularx}{1.2\textwidth}{ccccc}
            \hline
            & \begin{tabular}[c]{@{}c@{}}Topology\\ Optimization\end{tabular} & \begin{tabular}[c]{@{}c@{}}Parametric\\ Shape Optimization\end{tabular} & \begin{tabular}[c]{@{}c@{}}DGM-based\\ Shape Optimization\end{tabular} & Proposed Method\\ \hline
            Parameterization& Automatic
            & Manual
            & Automatic
            & Automatic
            \\ \hline
            Interpolation
            & Infeasible
            & Infeasible
            & Possible
            & Possible
            \\ \hline
            \begin{tabular}[c]{@{}c@{}}Convergent\\ possibility\end{tabular}
            & Bad
            & Good
            & Good
            & Good
            \\ \hline
            Training data
            & None
            & None
            & Large
            & \begin{tabular}[c]{@{}c@{}}Small\\ (minimum 2)\end{tabular}
            \\ \hline
        \end{tabularx}
    \end{adjustbox}
    }
\end{table}

{In summary, Section~\ref{method} is concisely represented by Table~\ref{tab:comparison}. Although DGM-based shape optimization successfully addresses the limitations of traditional topology optimization and parametric shape optimization methods, it still suffers from a critical drawback: requiring an extensive training dataset. This study addresses this issue by proposing a robust model that delivers strong performance even with limited datasets, making it particularly suitable for shape optimization applications. Detailed descriptions and explanations are provided in Section~\ref{sec:method}.}

\subsection{Deep Learning for 3D Data}

\begin{figure}[tbh]
    \centerline{\includegraphics[width=1.0\textwidth]{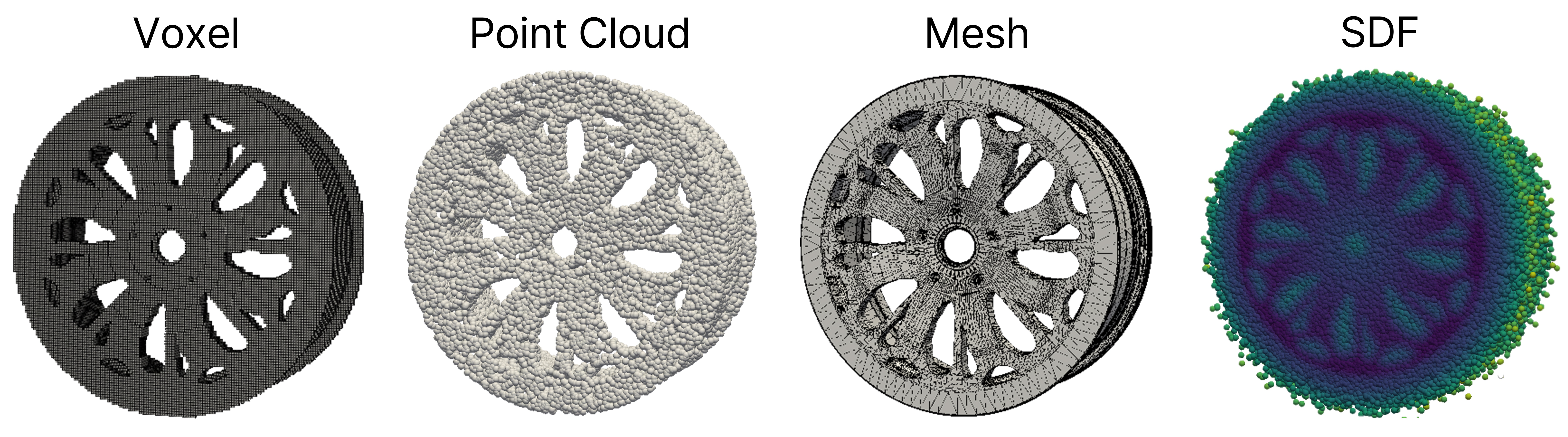}}
    \caption{{Four common 3D data representations}}
    \label{fig:data_representation}
\end{figure}

\subsubsection{3D Data Representation}
Deep learning with 3D data representation can vary depending on the specific problem being addressed. Representation for data-driven 3D deep learning can be broadly categorized into voxel, point cloud and mesh representation, as shown in Fig.~\ref{fig:data_representation} \unskip~\cite{regenwetter2022deep}.

First, voxel is a representation method that discretizes 3D volume into a grid. This is advantageous because it is similar to the representation of images, making it relatively easy to apply to traditional neural networks, such as convolutional neural networks \unskip~\cite{maturana2015voxnet}. However, using the voxel approach to represent 3D shapes can lead to memory issues. {Furthermore, representing a 3D shape as a voxel can result in relatively unnecessary areas, leading to higher computational costs.} Voxel representation is heavily influenced by resolution, which can make the shapes relatively less detailed and blunt.

Point cloud is a prominent representation method among 3D representations, particularly in applications such as classification, retrieval and segmentation \unskip~\cite{qi2017pointnet}. It represents a 3D surface using 3D points, allowing for detailed and global shape representations. However, it has a significant drawback in that it cannot capture connectivity or topological information between individual points and cannot generate watertight meshes.

Finally, mesh is a representation method that connects vertices and edges between them to represent the surface of a 3D shape. This method allows for the representation of detailed shapes and is memory-efficient, making it a popular choice in generative models. However, changing the topology can be challenging and self-intersection issues can arise when deforming the mesh \unskip~\cite{huang2021arapreg}. As a result, it is not a suitable representation for mechanical design with various topological changes.

At present, these representation methods are not suitable for learning 3D shapes, which is why there is growing interest in representing data implicitly. Implicit functions involve constructing a continuous volumetric field and embedding shapes as iso-surfaces, with the signed distance function (SDF) being a prominent example. 
This can be expressed in terms of the coordinates ($x$) in space as the signed distance ($s$) to the surface, as shown in Eq.~\ref{eqn:sdf_formulation}. The SDF represents the shortest distance from a point to the shape's surface while indicating whether the point is inside or outside the shape based on its sign. Specifically, for points inside the shape, the distance is negative ($-$), for points outside, it is positive ($+$) and it is zero on the surface.

\begin{equation}
\label{eqn:sdf_formulation}
    SDF(x) = s : x \in \mathbb{R}^3, s \in \mathbb{R}
\end{equation}

The surface of a shape represented by SDF can be expressed as the iso-surface of $SDF(\cdot) = 0$ and this can be converted into a discretized surface using algorithms such as marching cubes \unskip~\cite{lorensen1998marching}.

\subsubsection{Implicit Neural Representation}

Implicit representations such as SDF enable implicit learning, where coordinates serve as inputs to an AI model and predict the distance value corresponding to those coordinates. Since this type of learning does not rely on discretized data; thus, it theoretically enables exporting 3D shapes with infinite resolution and handling topological changes easily. Implicit learning can densely capture important parts of actual shapes, facilitating the generation of feasible and novel shapes. Starting with DeepSDF, the first to propose representing shapes as SDFs and learning them implicitly, several models have been introduced for shape manipulation. These include DualSDF, which adds a coarse network for shape manipulation and A-SDF, which disentangles the latent space for articulated objects \unskip~\cite{hao2020dualsdf, mu2021sdf}. All of these models feature an auto decoder (AD) structure without an encoder. Unlike many generative model studies that reduce data dimension within a bottleneck structure, these models do not have a separate encoder for data compression. Instead, the AD inputs both coordinates and latent code into the model and backpropagation optimizes latent code within the model.

Encoder decoder (ED) models that utilize encoders such as PointNet to compress data exist \unskip~\cite{qi2017pointnet}. These structures create latent codes through an encoder, which has been used to handle implicit data. The work by Peng et al. proposed a methodology for extracting both global and local features from point clouds, enabling the reconstruction of large 3D scenes \unskip~\cite{peng2020convolutional}. The paper introduced the idea of encoding point clouds into two-dimensional planes or 3D volumes, transforming unstructured point cloud data into structured latent space. {This methodology allows for assigning different latent codes to individual points, enabling learning more detailed features in shapes. They employed a U-Net for decoding. By using skip connections in the U-Net architecture, they could simultaneously learn both local and global features, significantly improving the accuracy of occupancy probability for each point. The derived latent vector and coordinates are then fed into an SDF network to predict the distance.}

\section{Methodology}
\label{sec:method}

\begin{figure}[tbh]
    \centerline{\includegraphics[width=0.9\textwidth]{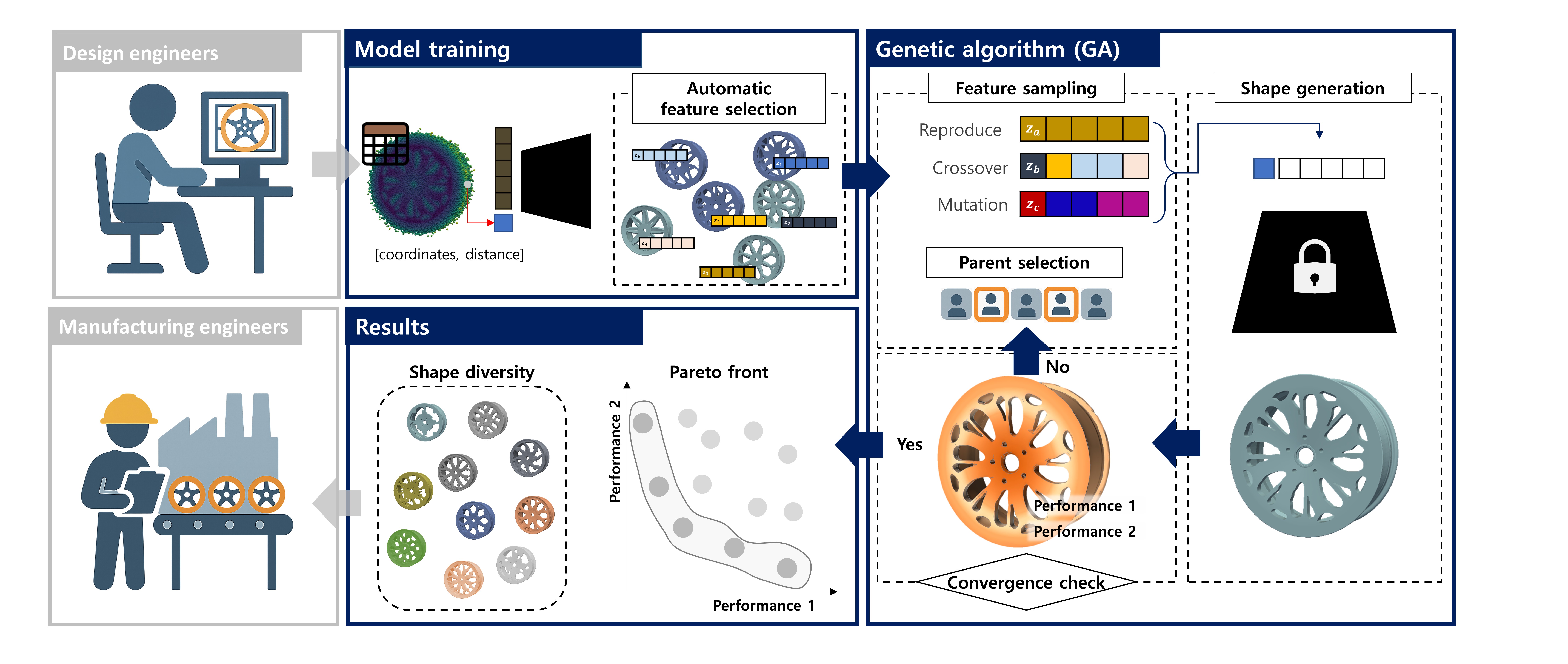}}
    \caption{{Overview of the proposed optimization }}
    \label{fig:framework}
\end{figure}

{The overall framework of this study is illustrated in Fig.~\ref{fig:framework}, which visually summarizes each step of the process, from data collection and preprocessing to model training, feature sampling and shape generation.}

\noindent \textbf{Data Collection:}

{In the initial phase, a limited dataset, comprising 30-wheel and 30-car data from both topology optimization and parametric design methods, was carefully selected to demonstrate that optimization can be effective even with limited data \unskip~\cite{chang2015shapenet, yoo2021integrating}. Preprocessing then ensured data quality through checks for mesh flips and watertightness \unskip~\cite{huang2018robust}, followed by centering, normalization and sampling surface-adjacent points with corresponding SDF extraction.}

\noindent \textbf{Model Training \& Automatic Feature Selection:}

{Preprocessed data, consisting of spatial coordinates and corresponding signed distances, enabled the model to learn their relationships. During training, latent codes were optimized alongside the model’s weights to enhance learning. To robustly capture fine details within a limited dataset, the implicit neural representation was improved by incorporating positional encoding for high-frequency information and employing Lipschitz regularization to build a meaningful latent space. Upon training completion, the learned decoder successfully reconstructed non-parametric data and generated a latent space that effectively vectorized shape information for the extraction of geometric features.}

\noindent \textbf{Feature Sampling:}

\begin{table}[tbh]
\centering

{
\begin{tabular}{cc}
\hline
Operator         & Used                       \\ \hline
Parent selection & Tournament selection       \\ \hline
Crossover        & Simulated binary crossover \\ \hline
Mutation         & Polynomial mutation        \\ \hline
\end{tabular}
\caption{{Types of operators used}}
\label{tab:ga_operators}
}
\end{table}

{With the training phase complete, the decoder was frozen and feature sampling was conducted from the continuous latent space. A genetic algorithm (GA) was employed as the optimization method \unskip~\cite{holland1992adaptation}, wherein a population is evolved over successive generations through key operators, parent selection, simulated binary crossover and polynomial mutation, to search for optimal solutions. Table~\ref{tab:ga_operators} summarizes the specific operators used. In GA, individuals with higher fitness are selected as parents, combined via crossover and then undergo mutation to maintain diversity and reduce the risk of converging to suboptimal solutions. This approach is well-suited for complex optimization challenges, enabling AI-based shape optimization by refining the latent code fed into the decoder. In particular, the NSGA-2 algorithm employs crowding distance calculations to explore a broader range of solutions and uncover diverse shape designs \unskip~\cite{deb2002fast, cunningham2020sparsity}.}

\noindent \textbf{Shape Generation \& Simulation:}

{Using latent codes produced by the optimization algorithm, the decoder generated new shapes by producing a SDF. This SDF was then converted into a 3D triangular mesh using the marching cubes algorithm \unskip~\cite{lorensen1998marching}.}

{The performance of the generated shapes was evaluated via simulations. FEM simulations were conducted using Altair Inspire, while CFD simulations were performed using OpenFOAM \unskip~\cite{altair2024accelerate, jasak2007openfoam}. Given the multi-objective nature of the optimization, two performance metrics with inherent trade-offs were selected: stiffness and mass for the wheel data and the drag coefficient ($C_D$) and lift coefficient ($C_L$) for the vehicle data.}

\noindent \textbf{Convergence Check:}

{Convergence was determined by evaluating performance metrics across a predefined maximum number of generations. If convergence wasn't achieved, the process returned to feature selection to create a new population for the next generation. If convergence was reached, the algorithm was terminated.}

\subsection{Network Architecture}

{Mechanical design products have boundary conditions, such as load and fixed conditions, resulting in minimal differences among the data. Therefore, it is important for generative models of mechanical design domains to learn the features of each dataset without significant differences between them.}

{Training on such a dataset with a standard multi-layer perceptron (MLP) may be problematic for two reasons. First, an MLP might fail to capture the unique features of each data sample, which is critical given that many mechanical components exhibit significant performance variations due to subtle differences. Second, the limited number of available samples within the same class poses a challenge. Hence, the AI model must accurately learn the features of the dataset even with sparse data while developing a meaningful latent space. Positional encoding and Lipschitz regularization terms have been introduced to address these issues.}

{Positional encoding, as outlined by Mildenhall et al. and Sitzmann et al., is a powerful technique that projects input coordinates into a high-dimensional Fourier space through sinusoidal transformations, as represented in Eq.~\ref{eqn:pe} \unskip~\cite{mildenhall2021nerf,sitzmann2020implicit}:}

\begin{equation}
\label{eqn:pe}
\gamma (p) = (sin(2^0 \pi p), cos(2^0 \pi p),..., sin(2^{L-1} \pi p), cos(2^{L-1} \pi p))
\end{equation}

$p$ represents the input coordinates and $L$  specifies the frequency levels in the encoding. This method allows the model to capture high-frequency information effectively before passing it to MLP.

{By encoding position information across a range of frequencies, the model is able to learn various spatial features and distinguish subtle variations within the defined space. Such granularity is advantageous for learning detailed SDFs, enabling the model to capture fine details in spatially complex data.}

{Furthermore, these studies have confirmed that projecting inputs into a higher-dimensional space using high-frequency functions, such as Fourier transformations, is highly effective for encoding complex information. This approach enhances model expressiveness and precision in capturing details.}

As presented by Liu et al., the Lipschitz regularization term smooths the latent space by constraining the magnitude of each layer’s weight matrix \unskip~\cite{liu2022learning}. This regularization is applied to each layer’s weight matrix within the MLP, as represented by the layer equation {$y = ReLU(W_ix + b_i)$}.

\begin{equation}
\label{eqn:lip_mlp}
{y = ReLU(\hat{W_i}x + b_i)}, \hat{W_i} = normalization(W_i, ln(1+e^{k_i}))
\end{equation}

{For all $l$ layers of the decoder, the Lipschitz layer normalizes the weight matrix $W_i$ of each $i$-th linear layer using trainable Lipschitz bounds $k_i$, as shown in Eq.~\ref{eqn:lip_mlp}.}

\begin{equation}
\label{eqn:lip_loss}
    loss_{Lipschitz} = \prod_{l}^{i = 1}ln(1+e^{k_i})
\end{equation}

{The overall Lipschitz bound for the network is determined by the product of the individual Lipschitz constants $k_i$ assigned to each layer. As shown in Eq.~\ref{eqn:lip_loss}, a loss function can enforce this constraint, stabilizing the model by preventing weight values from exceeding a set threshold at each layer. This regularization limits the model’s response to input data, reducing excessive sensitivity and enhancing overall model stability.}

{This method is particularly beneficial for synthesis tasks with limited data, as it confines the model’s output to a compact latent space, allowing a comprehensive representation of training data and effective pattern learning. Within the broader framework of limited-data optimization, Lipschitz regularization plays a pivotal role in fostering robust learning and ensuring reliable performance.}

\begin{figure}[h!]
    \centering
    \begin{subfigure}[tbh]{0.8\textwidth}
        \includegraphics[width=\textwidth]{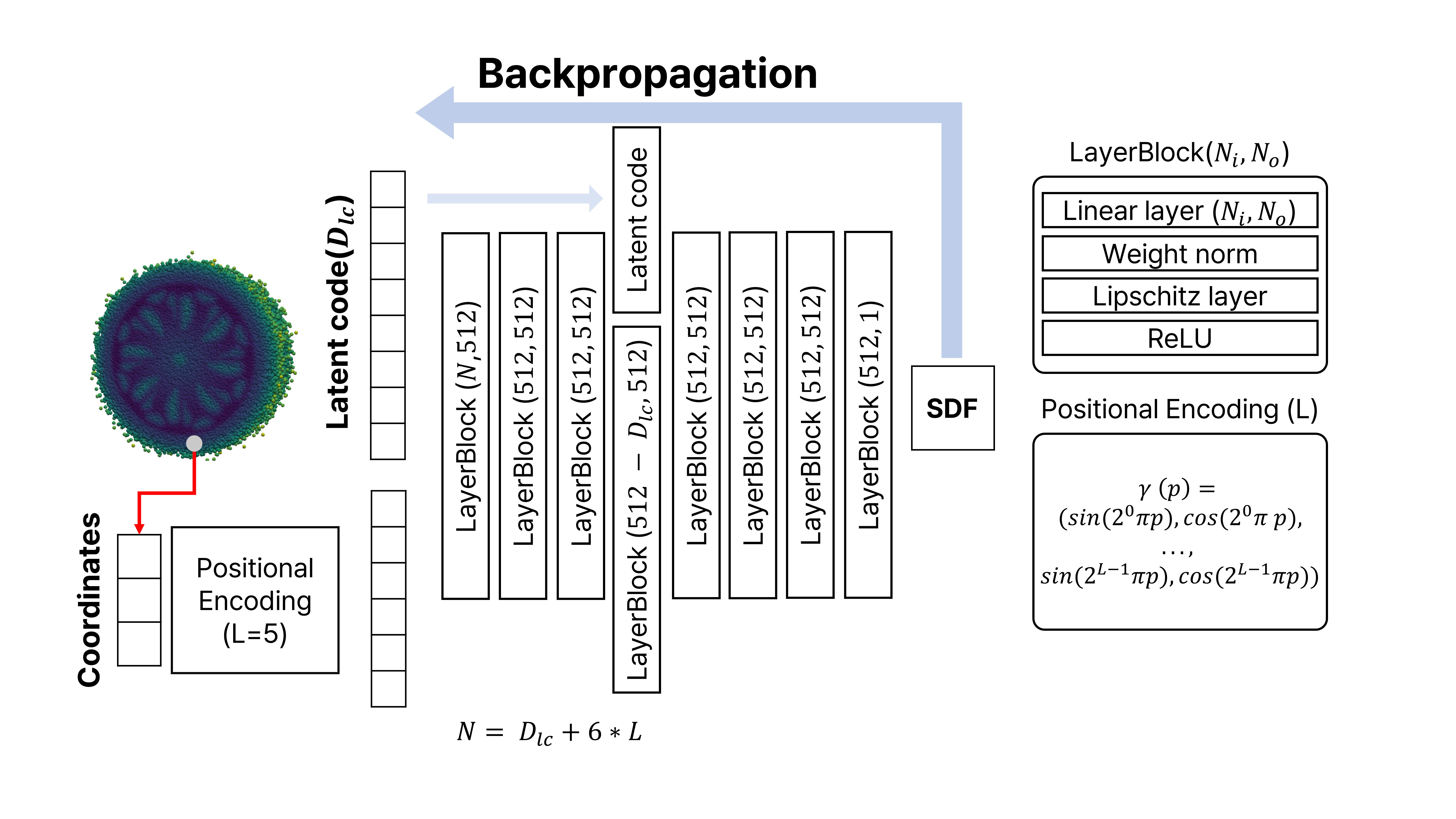}
        \caption{Auto decoder}
        \label{fig:ad-model}
    \end{subfigure}
    \begin{subfigure}[tbh]{0.8\textwidth}
            \includegraphics[width=\textwidth]{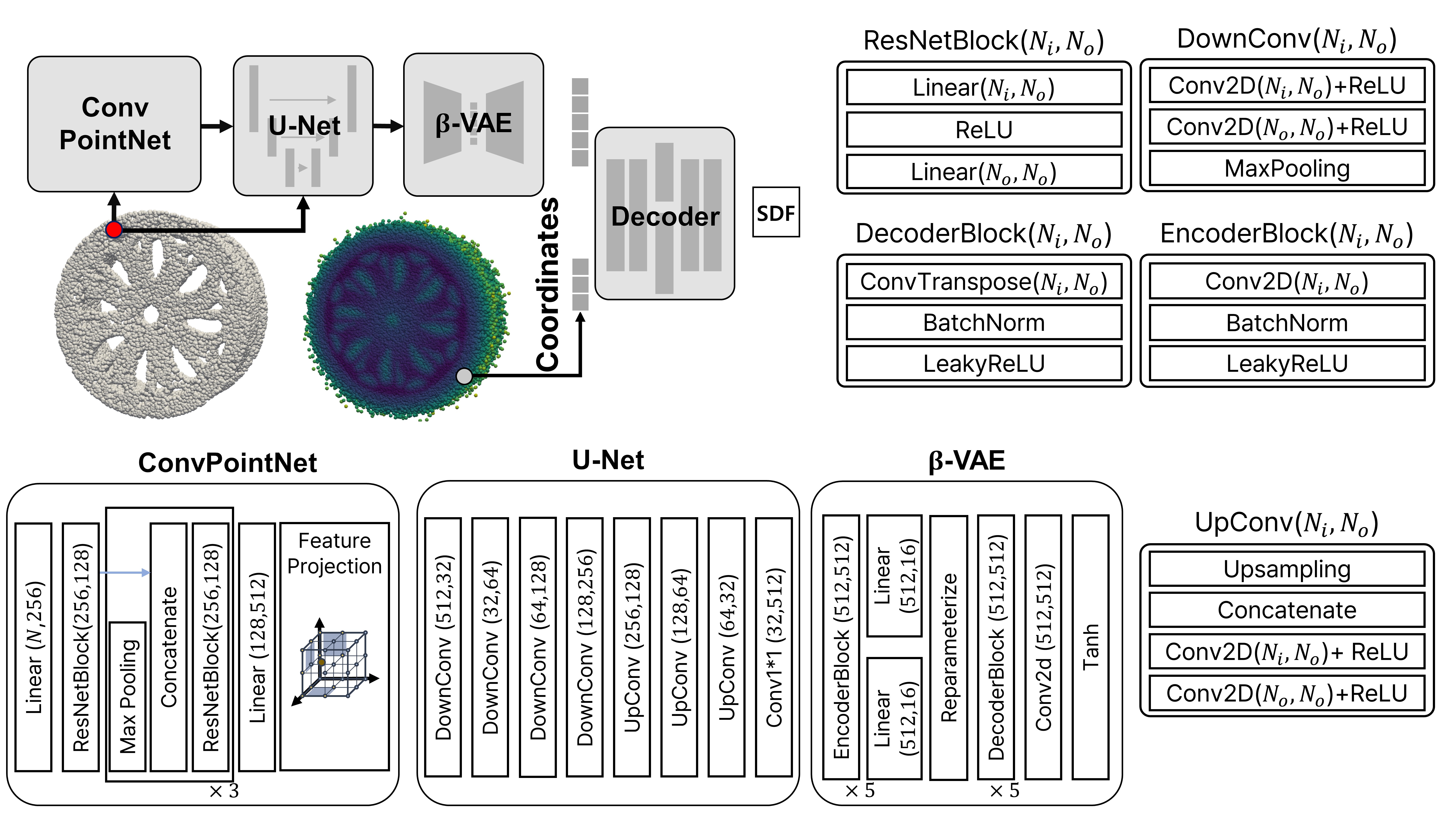}
            \caption{{Encoder decoder}}
            \label{fig:ed-model}
    \end{subfigure}
    \caption{Model architectures}
    \label{fig:model_architecture}
\end{figure}

Two architectures were evaluated: an AD structure without an encoder and an ED structure that integrates a PointNet-based encoder ($q_{\phi}$), as shown in Fig.~\ref{fig:model_architecture} \unskip~\cite{qi2017pointnet, peng2020convolutional}. This encoder inputs point cloud data ($\pi_{pcd}$) and extracts latent features. Both models are trained using a latent space regularization term and a truncated L1 loss to measure discrepancies between predicted and true SDF values. The truncated L1 loss, a modification of standard L1 loss, truncates both positive and negative outliers to limit extreme deviations while preserving sensitivity within a specified range ($\delta$), as illustrated in Eq.~\ref{eqn:l1loss}. {In experiments, $\delta$ was set to 0.1 and $\beta$ was set to 0.01.}

\begin{equation}
\label{eqn:l1loss}
    L_{clip}(d_{pred},d_{gt}) = \left\{\begin{matrix}
 max(d_{pred},-\delta) + \delta& d_{gt} < -\delta, \\
 |d_{pred}-d_{gt}|& -\delta<d_{gt}<\delta, \\
 \delta  - min(d_{pred},\delta)& d_{gt}>\delta.  \\
\end{matrix}\right.
\end{equation}

\begin{equation}
\label{eqn:loss_ad}
    L_{AD} =  L_{clip}(f_{\theta}(x,z),d_{gt}) + ||z||^2 + {w_{AD}}loss_{Lipschitz}
\end{equation}

\begin{equation}
\label{eqn:loss_ed}
    L_{ED} =  L_{clip}(f_{\theta}(x,z),d_{gt}) + \beta(D_{KL}(q_{\phi}(z|\pi_{pcd})||f_{\theta}(z))) + {w_{ED}}loss_{Lipschitz}
\end{equation}

The overall loss function comprises multiple components, as defined in Eqs.~\ref{eqn:loss_ad} and \ref{eqn:loss_ed}.

Let $f_{\theta}$ denote the AI model implicitly learning SDF. Here, $z_{i}$ represents the latent vector embedding each shape, while $x$ and $d_{gt}$ indicate the spatial coordinates and the corresponding ground truth distance. {The $\beta$-VAE was trained by computing the KL divergence, $D_{KL}$, between the learned latent distribution and the prior. To align the scale with $L_{\text{clip}}$ and $D_{KL}$, weights of $w_{AD} = 1\times10^{-7}$ for the AD model and $w_{ED} = 1\times10  ^{-5}$ for the ED model were applied to the Lipschitz loss term.}

{An AD structure without an encoder was adopted; the rationale behind this design choice is detailed below:}

\begin{enumerate}[label=\arabic*)]
    \item Reconstruction performance
    
    {Reconstruction performance was compared for the AD and ED models using chamfer distance (CD), minimum matching distance (MMD) and coverage (COV) metrics, as introduced by Yang et al. \unskip~\cite{yang2019pointflow}. These metrics serve specific purposes. The formulas for CD, MMD and COV are as Eqs.~\ref{eqn:cd}, \ref{eqn:mmd} and \ref{eqn:cov}, respectively. 
    Let $P_{g}$ and $P_{r}$ denote the sets of generated and reference point clouds, respectively, with $|P_{r}| = |P_{g}|$. The distance $D(\cdot, \cdot)$ between two point clouds is computed using either CD or earth mover's distance.} CD measures similarity, MMD measures quality and COV measures diversity. COV serves as an indicator for mode collapse. 
    \begin{equation}
    \label{eqn:cd}
        CD(P_g, P_r) = \sum_{g \in P_g} \min\limits_{r \in P_r} || g - r||^2 + \sum_{r \in P_r} \min\limits_{g \in P_g} ||r- g||^2 
    \end{equation}
    
    \begin{equation}
    \label{eqn:mmd}
        MMD(P_{g}, P_{r}) = \frac{1}{|P_{r}|} \sum_{r \in P_{r}} \min\limits_{g \in P_{g}} D(g,r)
    \end{equation}
    
    \begin{equation}
    \label{eqn:cov}
        COV(P_{g}, P_{r}) = \frac{|\{argmin_{r\in P_{r}}D(g,r)| g \in P_{g}\}|}{|P_{r}|}
    \end{equation}

    {To evaluate both models, experiments were performed on 30 car shapes from ShapeNet and car wheels generated via deep generative models and parametric design \unskip~\cite{chang2015shapenet, yoo2021integrating}. Reconstructions were obtained using the trained models, with 20,000 sampled points used to compute metrics. As shown in Table~\ref{tab:train-results}, the AD model consistently outperformed the ED model for both car and wheel shapes, with the CD (mean) indicating that the AD handles subtle variations effectively. Although the CD (mean) favored the AD, the CD (median) was lower for the ED, suggesting that mode collapse impacted the ED’s performance, particularly in smaller latent spaces. Moreover, as shown in Fig.~\ref{fig:data_recon_comparison}, increasing the latent dimension in the AD led to difficulties in reconstructing thin structures (such as wheel rims), while a smaller latent dimension allowed the model to capture fine features more comprehensively, preserving a diverse and meaningful latent space.}

    \begin{table}[tbh]
        \caption{Training results}
        \label{tab:train-results}
        \addtocounter{table}{-1}
        \begin{adjustbox}{minipage=1.3\textwidth, center}
            \begin{tabularx}{1.1\textwidth}{cccccccc}
                \hline
                Data& Structure& \begin{tabular}[c]{@{}c@{}}Latent\\ dimension\end{tabular} & Time& \begin{tabular}[c]{@{}c@{}}CD (↓) \\ (mean, ×$10^3$)\end{tabular} & \begin{tabular}[c]{@{}c@{}}CD (↓)\\ (median, ×$10^3$)\end{tabular} & \begin{tabular}[c]{@{}c@{}}MMD (↓)\\ ×$10^3$\end{tabular} & COV (↑) \\ \hline
                \multirow{6}{*}{Car}   & \multirow{4}{*}{\begin{tabular}[c]{@{}c@{}}Auto\\ decoder\end{tabular}}     & 2& \multirow{4}{*}{4h} & 0.227& 0.189& 0.222& \textbf{0.967}   \\ \cline{3-3} \cline{5-8} 
                                       && 5&& 0.232& 0.120& \textbf{0.208}& \textbf{0.967}   \\ \cline{3-3} \cline{5-8} 
                                       && 10&& \textbf{0.226}& 0.151& 0.227& 0.933   \\ \cline{3-3} \cline{5-8} 
                                       && 128&& 0.517& 0.352& 0.503& 0.833   \\ \cline{2-8} 
                                       & \multirow{2}{*}{\begin{tabular}[c]{@{}c@{}}Encoder\\decoder\end{tabular}} & 6& \multirow{2}{*}{2d} & 3.203& 2.762& 2.369& 0.167   \\ \cline{3-3} \cline{5-8} 
                                       && 384&& 0.535& \textbf{0.098}& 0.386& 0.633   \\ \hline
                \multirow{6}{*}{Wheel} & \multirow{4}{*}{\begin{tabular}[c]{@{}c@{}}Auto\\ decoder\end{tabular}}& 2& \multirow{4}{*}{4h} & 0.113& 0.114& 0.113& \textbf{1}\\ \cline{3-3} \cline{5-8} 
                                       && 5&& \textbf{0.106}& \textbf{0.106}& \textbf{0.106}& \textbf{1}\\ \cline{3-3} \cline{5-8} 
                                       && 10&& 0.128& 0.128& 0.128& \textbf{1}\\ \cline{3-3} \cline{5-8} 
                                       && 128&& 3.390& 5.740& 0.311& 0.533   \\ \cline{2-8} 
                                       & \multirow{2}{*}{\begin{tabular}[c]{@{}c@{}}Encoder\\decoder\end{tabular}} & 6& \multirow{2}{*}{2d} & 0.373& 0.359& 0.358& 0.033   \\ \cline{3-3} \cline{5-8} 
                                       && 384&& 0.152& 0.129& 0.144& 0.633   \\ \hline
            \end{tabularx}
        \end{adjustbox}
    \end{table}
    
    \item Relatively short training time

    {The ED structure incorporates an encoder, resulting in 2.1 to 11.1 times more parameters than the AD structure, which contains 1,843,655 parameters. Both the AD and ED models were trained on 15,000 paired data points for SDF learning and the ED model additionally incorporated 4,096 point‑cloud samples into the training process.}
    
    The slower training speed of the ED model is primarily due to its complex processing rather than just a larger parameter count. {Comprising a $\beta$-VAE, PointNet and U-Net for global feature extraction \unskip~\cite{higgins2017beta}, the ED model performs additional operations like projecting point clouds onto two dimensions, increasing processing time. Beyond its computational burden, the model performs additional internal tasks, such as projecting point clouds into two-dimensional planes. These operations consume considerable processing time, significantly challenging the overall training speed.} Specifically, training the ED model on 30 samples required about 2 days on an Nvidia A100 GPU over 40,000 epochs, which is a significant drawback when rapid training is required. In contrast, the simplicity of the AD structure reduces computational overhead and training time dramatically; it requires only about 4 hours over 8,000 epochs, highlighting its efficiency for rapid deployment.

    \begin{figure}[h!]
        \centerline{\includegraphics[width=0.8\textwidth]{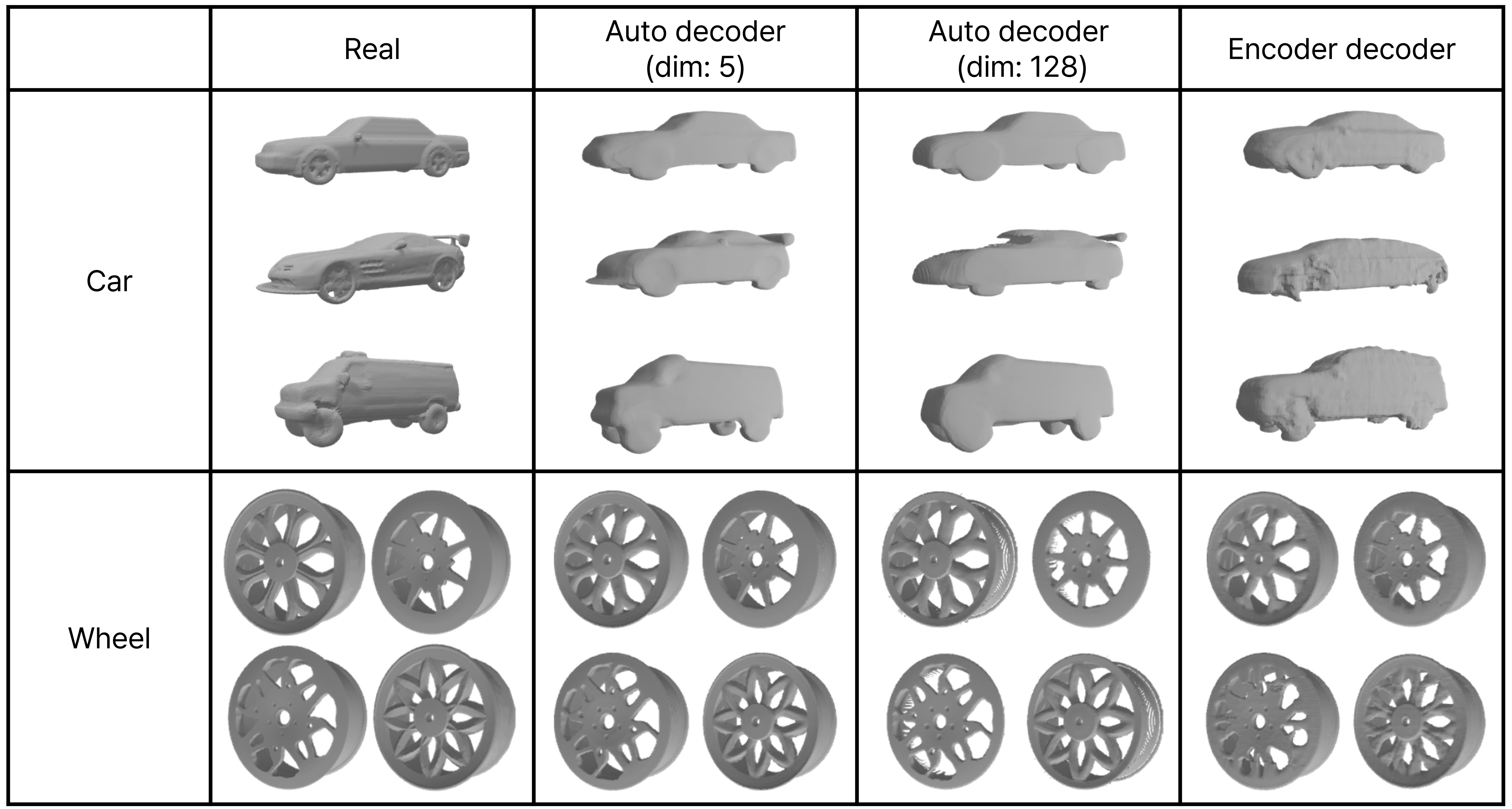}}
        \caption{Reconstructed 3D shapes for car and wheel datasets}
        \label{fig:data_recon_comparison}
    \end{figure}
    
\end{enumerate}

\pagebreak

\section{Experiments}
\label{sec:exp}

\begin{table}[tbh]
    \caption{Experiment descriptions}
    \label{tab:experiments-descriptions}
    \begin{adjustbox}{minipage=1.5\textwidth, center}
        \begin{tabularx}{1.5\textwidth}{cccccc}
            \hline
            Dataset& Objective& Experiment& \# of data & {\begin{tabular}[c]{@{}c@{}}Latent\\ Dimension\end{tabular}} &Use of Neural Network\\ \hline
            \multirow{2}{*}{Wheel}& \multirow{2}{*}{\begin{tabular}[c]{@{}c@{}}max: stiffness\\ min: mass\end{tabular}} & Data Synthesis  & 30 & {5} & Synthesize the other types of data\\ \cline{3-6} 
            && {\begin{tabular}[c]{@{}c@{}}Extremely \\ Limited Data\end{tabular}} & 2 & {1} & Explore shapes between 2 data\\ \hline
            \multirow{2}{*}{Car} & \multirow{2}{*}{min: $C_D, C_L$}& CFD Analysis    & 30 & {5} & \begin{tabular}[c]{@{}c@{}}Generate good-quality shapes\\ for engineering analysis\end{tabular} \\ \cline{3-6} 
            && {\begin{tabular}[c]{@{}c@{}}Extremely \\ Limited Data\end{tabular}} & 2 & {1} & Explore shapes between 2 data\\ \hline
        \end{tabularx}
    \end{adjustbox}
\end{table}

In this section, experiments were conducted to demonstrate the novelty and versatility of {the proposed} methodology. Particularly, this study utilizes data-driven methods to generate shapes with various features based on the training data. Multi-objective optimization was performed to leverage the ability to explore diverse shapes in {the proposed} approach. {The proposed} methodology's versatility was demonstrated through experiments conducted in the context of shape optimization across {distinct} data types, CFD-based shape optimization and optimization between two shapes. The experiment concerning the optimization between two shapes was conducted using car data and wheel data. A brief description of the experiments is provided in Table~\ref{tab:experiments-descriptions}. {Experiments conducted in Section~\ref{sec:wheel-exp30} and \ref{sec:car-exp30} were performed with a population size of 10 individuals per generation, iterating through a total of 20 generations. Similarly, the experimental settings described in Section~\ref{sec:exp2} employed a population size of 8 individuals per generation over 20 generations.}

\subsection{Shape Optimization across {Distinct} Data {Types}}
\label{sec:wheel-exp30}

\begin{figure}[tbh]
    \makebox[\textwidth][c]{                
                \begin{subfigure}[tbh]{0.5\textwidth}
                    \includegraphics[width=\textwidth]{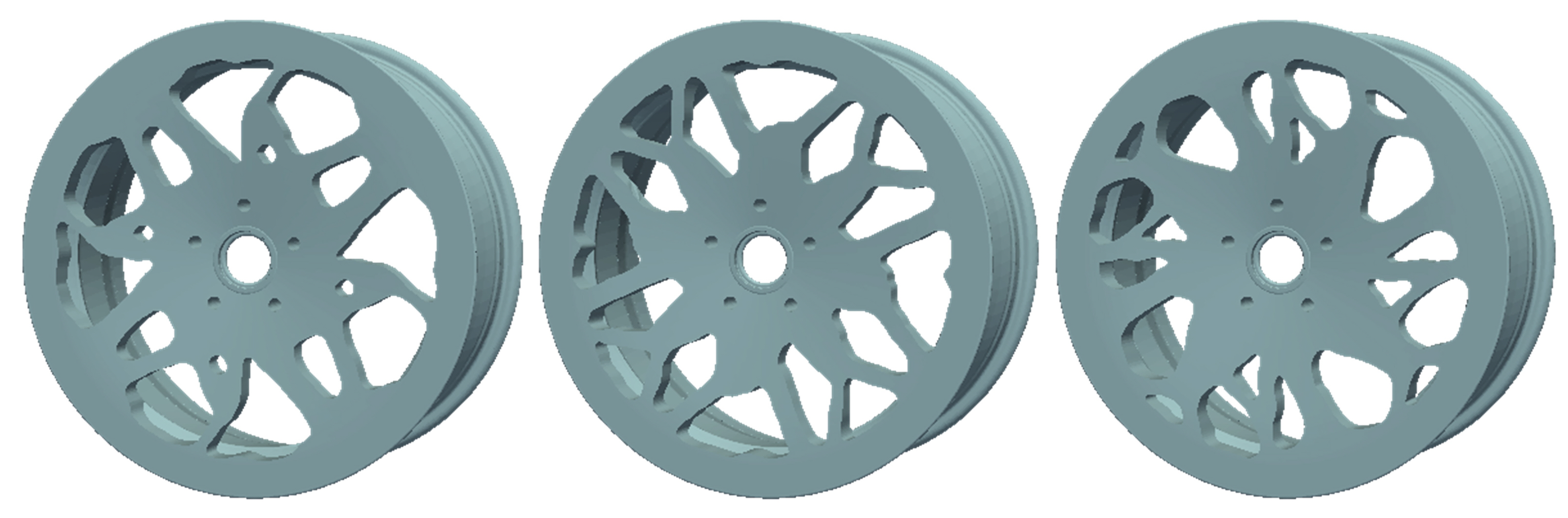}
                    \caption{Generative design data}
                    \label{fig:to-data}
                \end{subfigure}
                \begin{subfigure}[tbh]{0.5\textwidth}
                    \includegraphics[width=\textwidth]{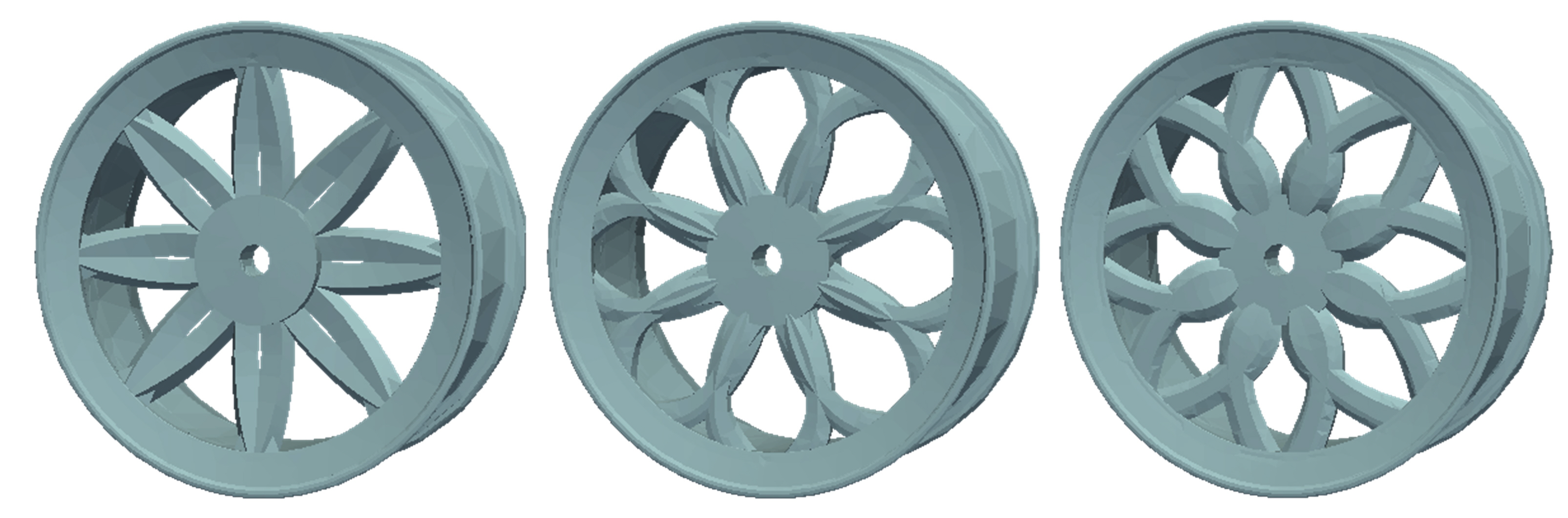}
                    \caption{Parametric design data}
                    \label{fig:pd-data}
                \end{subfigure}}
    \caption{{Different wheel design variations}}
    \label{fig:wheel_datasets}
\end{figure}

Traditional parametric shape optimization represents arbitrary shapes using a single set of parameters, potentially sacrificing some intrinsic degrees of freedom. Similarly, reducing the full dimensionality of 3D input shapes to a compressed latent space vector may result in some loss of information. {However, the learned latent space is designed to capture the most significant shape variations, which means it is more likely to preserve critical degrees of freedom across a broader range of shapes than conventional parametric representation techniques.}

As seen in Fig.~\ref{fig:wheel_datasets}, even when the same wheel is designed, data can be defined in a variety of ways. Defining various datasets as a single parametric set is impossible. As a result, no research on optimizing them has been conducted. The methodology of this study is based on data rather than parametric modeling, which is why it is not heavily dependent on the data generation process. {To demonstrate this condition, experiments were conducted using a dataset comprising 13 engineer-designed wheels and 17 wheels generated by a deep generative model \unskip~\cite{oh2019deep,yoo2021integrating}.} The selection process involved identifying wheel data with a spoke pattern composed of multiples of 4, allowing for a symmetrical shape achieved through a 90-degree rotation, considering the significant characteristic of cyclic symmetry in wheel data.
In this study, stiffness is obtained by analyzing the natural frequencies of the wheels through modal analysis using the free-free method \cite{yoo2021integrating}. The free-free method in modal analysis examines the natural frequencies and mode shapes of a structure in an unconstrained state, meaning the object has no fixed boundaries or supports. This is done by solving the eigenvalue problem derived from the system's equations of motion, which are based on its mass and stiffness properties. The natural frequencies correspond to the square roots of the eigenvalues, indicating the rates at which the structure vibrates naturally. This approach allows for an accurate assessment of the dynamic response of the structure without the influence of external constraints, which is crucial for predicting the behavior of components in real-world scenarios, especially in aerospace and mechanical applications.

Natural frequencies ($f$) play a fundamental role in determining a structure's stiffness ($k$), with a direct relationship described by Eq.~\ref{eqn:mode}, where $m$ represents the mass of the structure. This formula indicates that an increase in natural frequency corresponds to higher stiffness, whereas an increase in mass yields lower stiffness. Natural frequencies indicate the rates at which the structure vibrates naturally when disturbed. Stiffness measures a structure's resistance to deformation under an applied force, influencing how the structure responds to external loads and vibrations.

\begin{figure}[tbh]    \centerline{\includegraphics[width=1.0\textwidth]{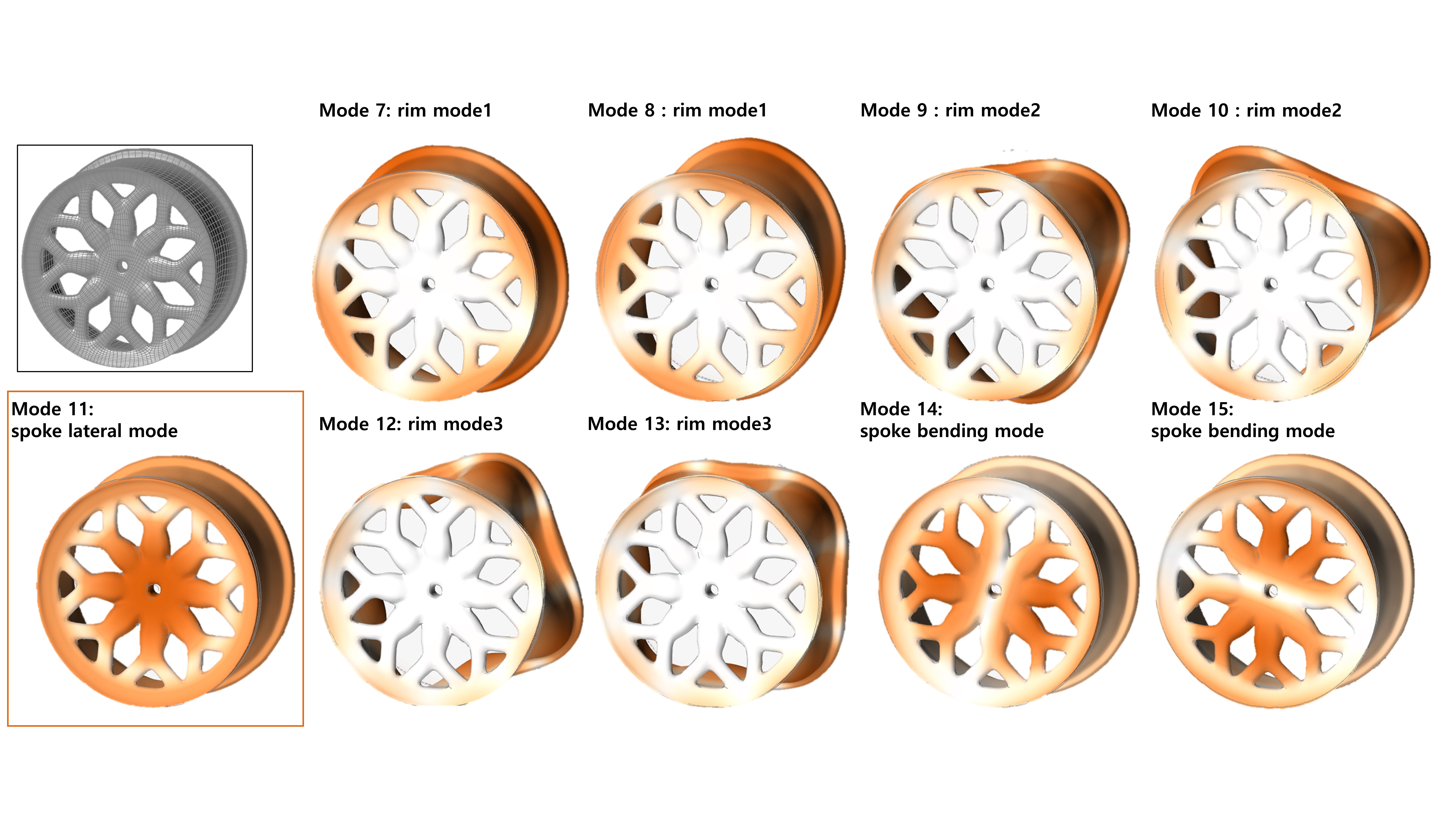}}
    \caption{{Modal analysis results: mode shapes across different modes}}
    \label{fig:mode-shapes}
\end{figure}

In this study, the stiffness was specifically calculated using the 11th mode, which is identified as the lateral mode of the spokes. The lateral mode involves vibrations that occur perpendicular to the wheel's plane, directly affecting the structural integrity and performance of the wheel design. (Fig.~\ref{fig:mode-shapes})

\begin{equation}
\label{eqn:mode}
    k = m(2 \pi f)^2
\end{equation}

\begin{figure}[h!]
    \centering
    \begin{subfigure}[tbh]{0.55\textwidth}

        \includegraphics[width=\textwidth]{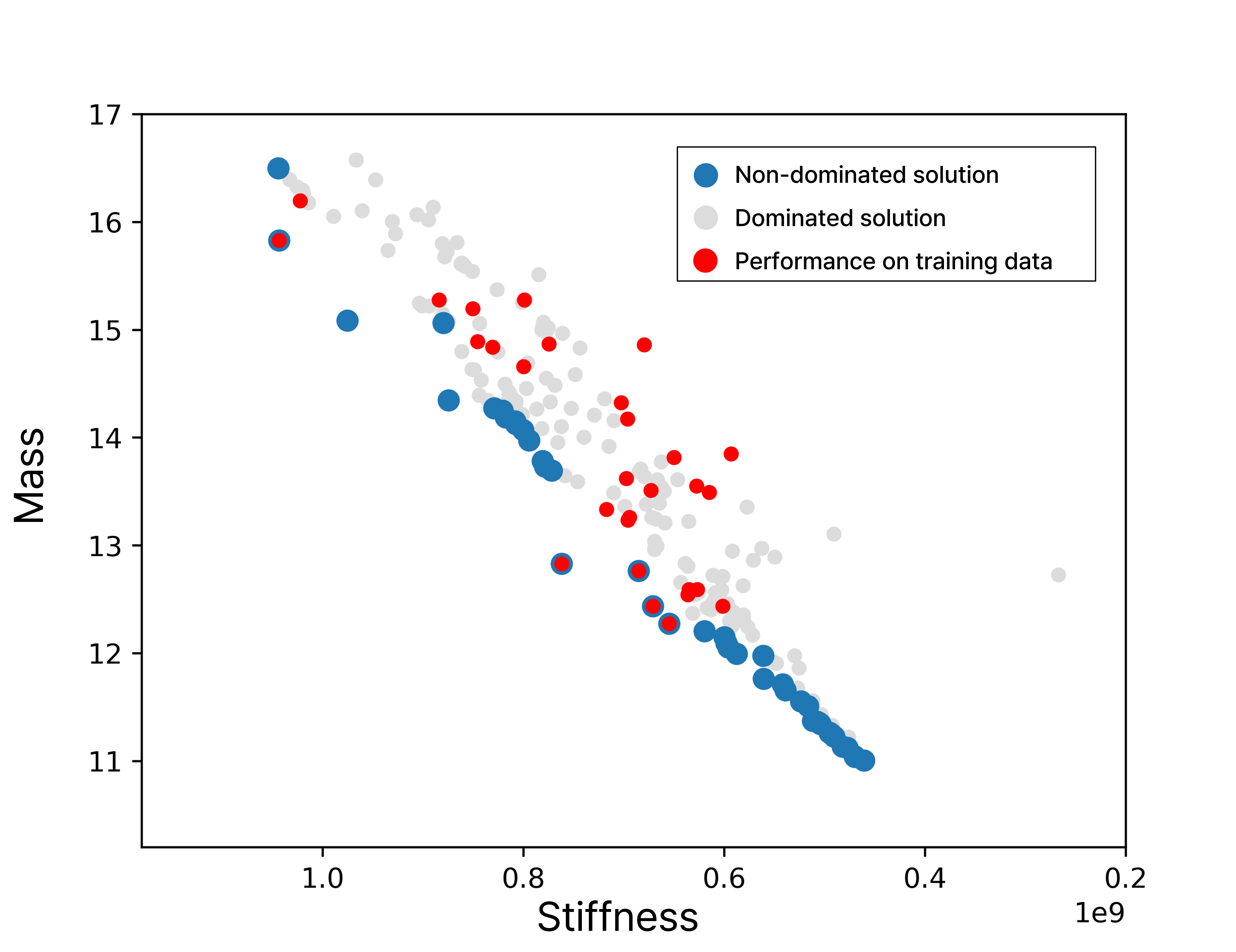}
        \caption{{Pareto front results using 30 wheel samples}}
        \label{fig:wheel30-pareto}
    \end{subfigure}
    \begin{subfigure}[tbh]{0.8\textwidth}
        \includegraphics[width=\textwidth]{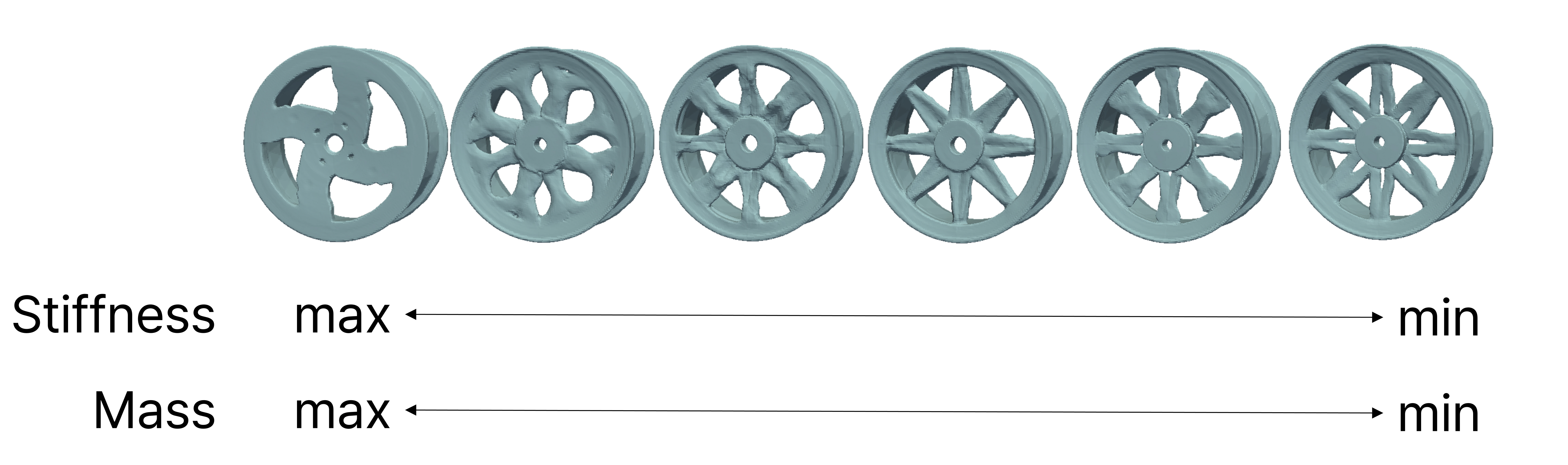}
        \caption{{Non-dominated solutions from 30 wheel samples}}
        \label{fig:wheel30-pareto-shapes}
    \end{subfigure}
    \caption{Optimization experiment result using two datasets}
    \label{fig:wheel30_experiment}
\end{figure}

{This experiment conducted multi-objective optimization with stiffness and mass as objectives. The results, presented in Fig.~\ref{fig:wheel30-pareto}, illustrate the evolution of shape designs positioned along the Pareto front across multiple generations. These outcomes demonstrate the proposed method's ability to continuously explore data within a low-dimensional latent space rather than sampling and optimizing disconnected regions of the data space. The method successfully identifies meaningful candidate designs around the Pareto front, even when compared to the training data. Furthermore, Fig.~\ref{fig:wheel30-pareto-shapes} reveals that the proposed approach, being fundamentally data-driven, enables realistic exploration within a design space closely related to existing shapes. Notably, the generated designs consistently exhibit high quality, irrespective of their origin from either parametric or generative design methods. The experiment thus demonstrates the capability of the proposed methodology to effectively explore and optimize within a continuous, low-dimensional data space rather than merely selecting solutions from isolated and discrete regions.}

\subsection{CFD-based Shape Optimization}
\label{sec:car-exp30}

\begin{figure}[h!]
    \centerline{\includegraphics[width=1.0\textwidth]{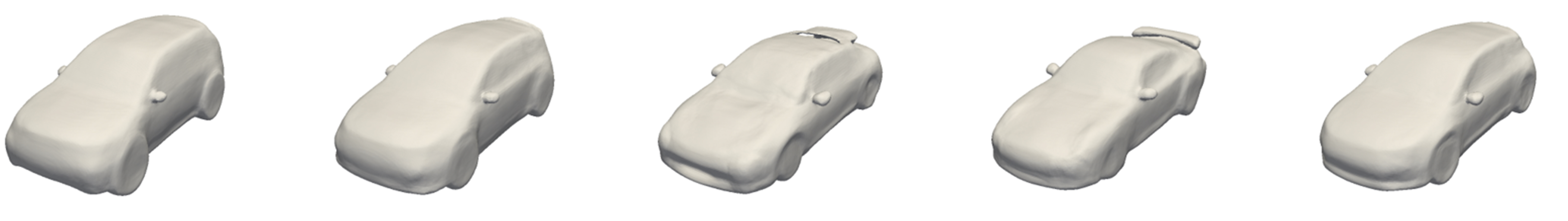}}
    \caption{Diverse car shape expression}
    \label{fig:car_interpolation}
\end{figure}

{Experiments analyzing car aerodynamics were conducted as part of a generative model-based shape optimization study. In this work, the OpenFOAM solver was set up for an incompressible Reynolds Averaged Navier-Stokes analysis using air. The simulation used SimpleFoam to capture the drag and lift forces generated as an external airflow, moving at 10 m/s, passes by the vehicle.}

{3D car models were sourced from the ShapeNet dataset, a widely used benchmark in 3D graphics. From thousands of models, those with clearly distinguished internal and external components and suitable mesh conditions for CFD analysis were selected. The study included various car categories, such as sedans, sports cars, cars with rear spoilers and sport utility vehicles.}

{Unlike traditional approaches where explicit parameterization is required to represent and compare topological changes, this data-driven methodology leverages generative models to directly capture the inherent shape features \unskip~\cite{chen2021padgan, regenwetter2022towards}. This allows for the representation and comparison of topological variations solely based on the input data, enabling the optimization process to preserve fine details that conventional parameterizations often miss.}

{Using the OpenFOAM simulation framework, a generative model was trained on a selection of 30 ShapeNet car models. Multi-objective optimization was performed to minimize the $C_D$ and the $C_L$.}

\begin{figure}[h!]
    \centering
    \begin{subfigure}[tbh]{0.55\textwidth}
        \includegraphics[width=\textwidth]{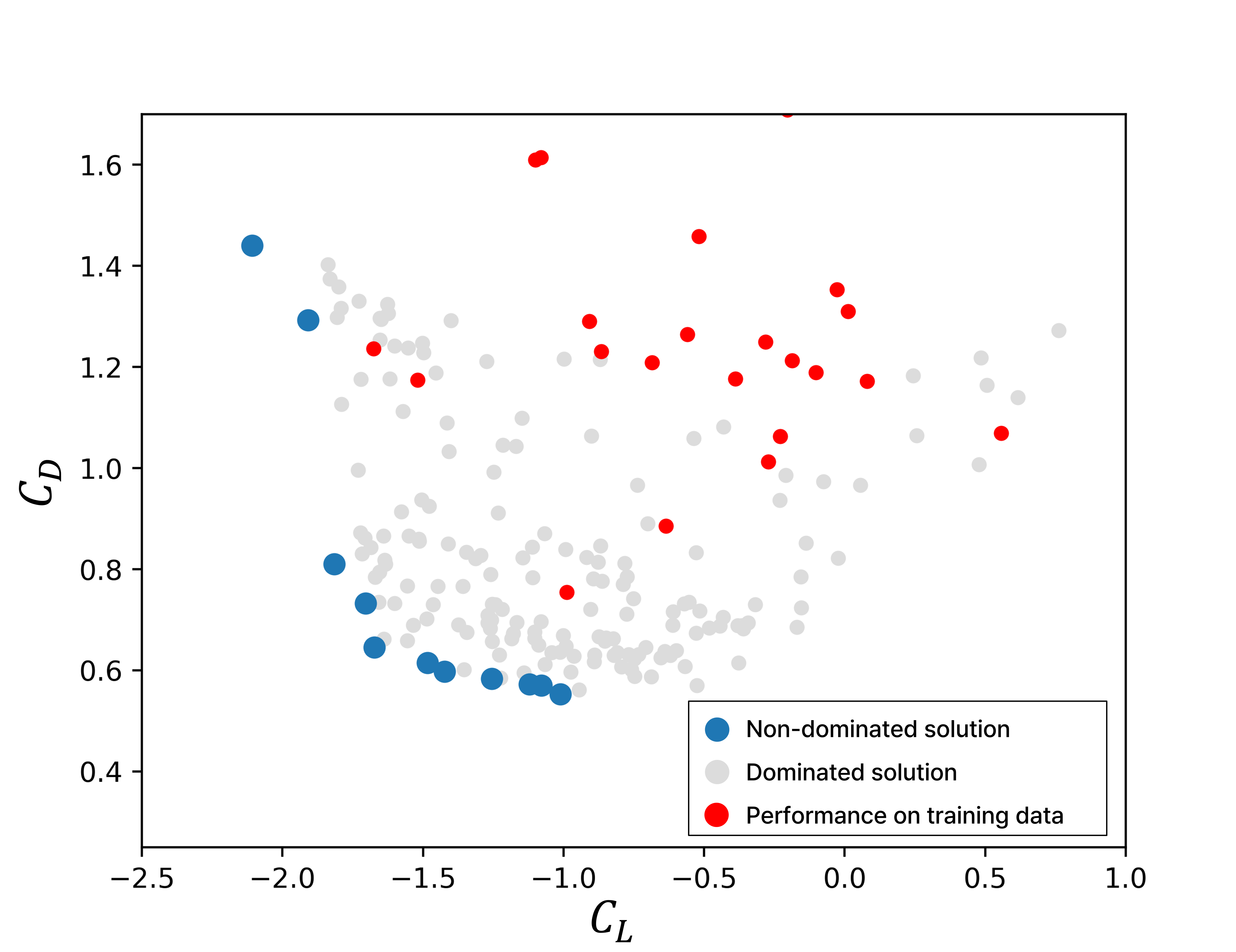}
        \caption{{Pareto front results using 30 car samples}}
        \label{fig:car30-pareto}

    \end{subfigure}
    \begin{subfigure}[tbh]{0.8\textwidth}
        \includegraphics[width=\textwidth]{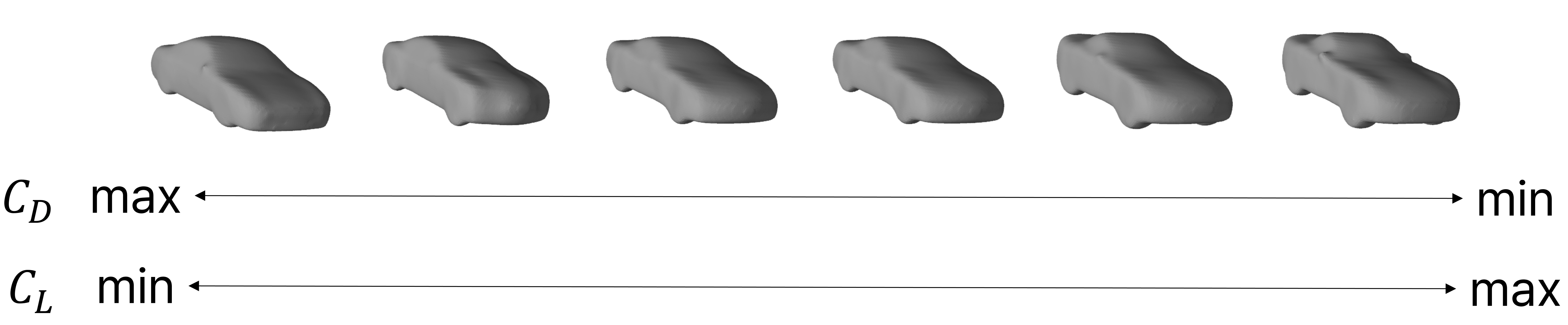}
        \caption{{Non-dominated solutions from 30 car samples}}
        \label{fig:car30-pareto-shapes}
    \end{subfigure}
    \caption{CFD-based car shape optimization results}
    \label{fig:car30_experiment}
\end{figure}

{The results are shown in Fig.~\ref{fig:car30-pareto}. As the generations progressed, improved designs emerged and gradually converged toward the Pareto front, following a trend similar to previous experiments. Notably, the generated candidates demonstrated superior aerodynamic performance compared to the training data. Furthermore, compared with the earlier wheel dataset study, variations in car shapes exhibited a more pronounced influence on the objective values, resulting in a broader distribution along the Pareto front. Fig.~\ref{fig:car30-pareto-shapes} further highlights the diversity of optimized shapes spanning various vehicle categories, from sports cars to sedans, illustrating the model's capability to explore and generate diverse Pareto-optimal designs under the given objectives.}

\pagebreak

\subsection{Optimization between Two {Shapes}}
\label{sec:exp2}

\begin{table}[tbh]
\centering
{
\begin{tabular}{cccc}
\hline
Data& \begin{tabular}[c]{@{}c@{}}CD (↓) \\ (mean, ×$10^3$)\end{tabular} & \begin{tabular}[c]{@{}c@{}}MMD (↓)\\ ×$10^3$\end{tabular} & COV (↑) \\ \hline
Car & 0.113   & 0.091 &1.000                       \\ \hline
Wheel & 0.073 & 0.113 &1.000                       \\ \hline
\end{tabular}
\caption{{Reconstruction results for 2 data}}
\label{tab:2data_recon}
}
\end{table}

Building on the foundational work described in Sections~\ref{sec:wheel-exp30} and \ref{sec:car-exp30}, this combined section examines the application of {the proposed} optimization framework with a {limited} dataset, using only two shapes for both wheels and cars. In contrast to the previous experiments, which employed a larger and more varied dataset of 30 designs, these experiments were designed to evaluate the methodology’s robustness and adaptability under extremely limited data availability. {Identical experimental conditions to those described in Sections~\ref{sec:wheel-exp30} and \ref{sec:car-exp30} were maintained to assess the optimization framework’s ability to derive meaningful design variations, with the only differences being a reduced number of data and a lower-dimensional latent space. The reconstruction accuracy obtained with this two‑sample setup is summarized in Table~\ref{tab:2data_recon}.}

\begin{figure}[tbh]
    \centering
    \begin{subfigure}[tbh]{1.0\textwidth}
        \makebox[\textwidth][c]{
                    \begin{subfigure}[tbh]{0.49\textwidth}
                        \includegraphics[height=5.0cm]{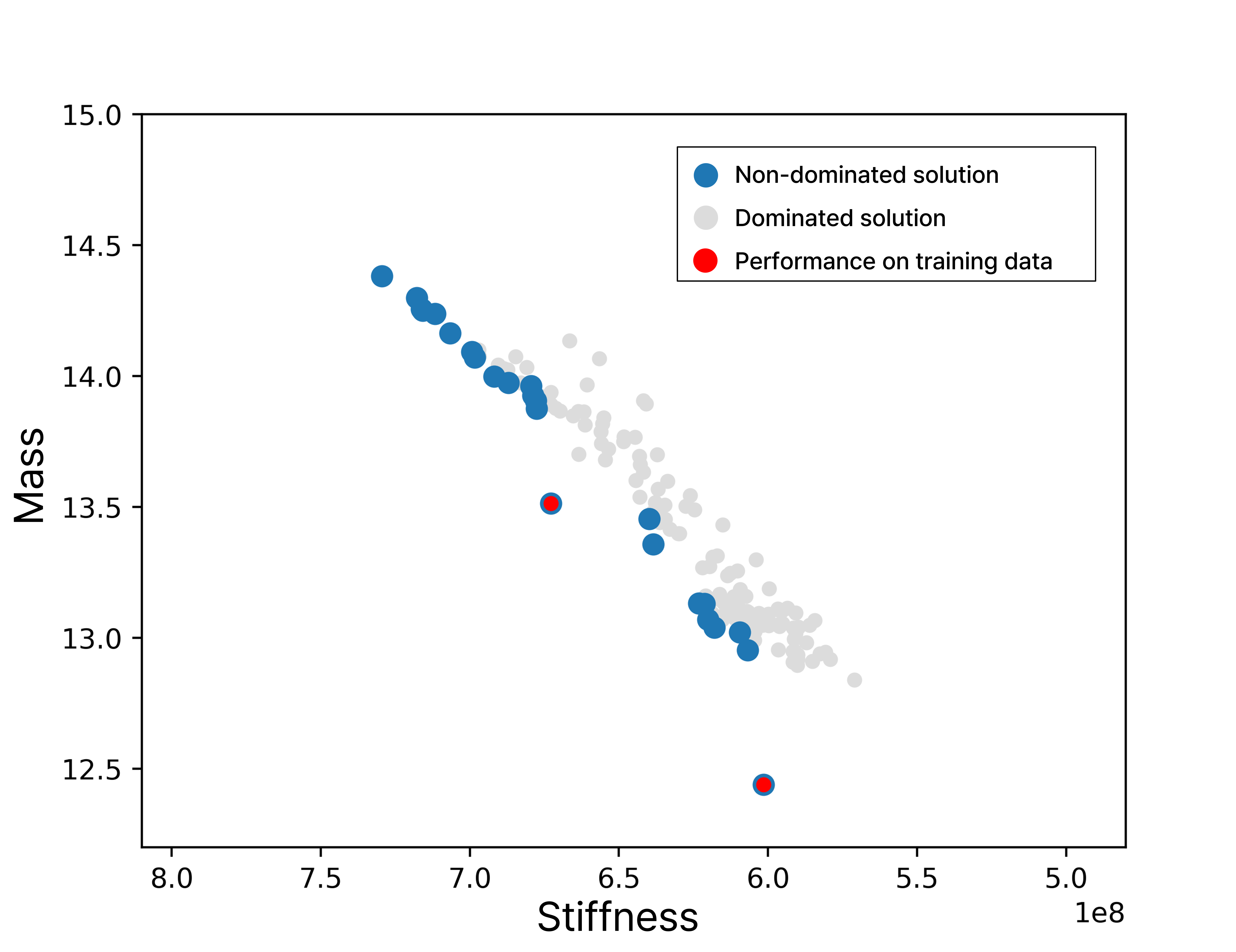}
                        \caption{{Pareto front results using two wheel samples}}
                        \label{fig:wheel2-pareto}
                    \end{subfigure}
                    \begin{subfigure}[tbh]{0.5\textwidth}
                        \includegraphics[height=5.0cm]{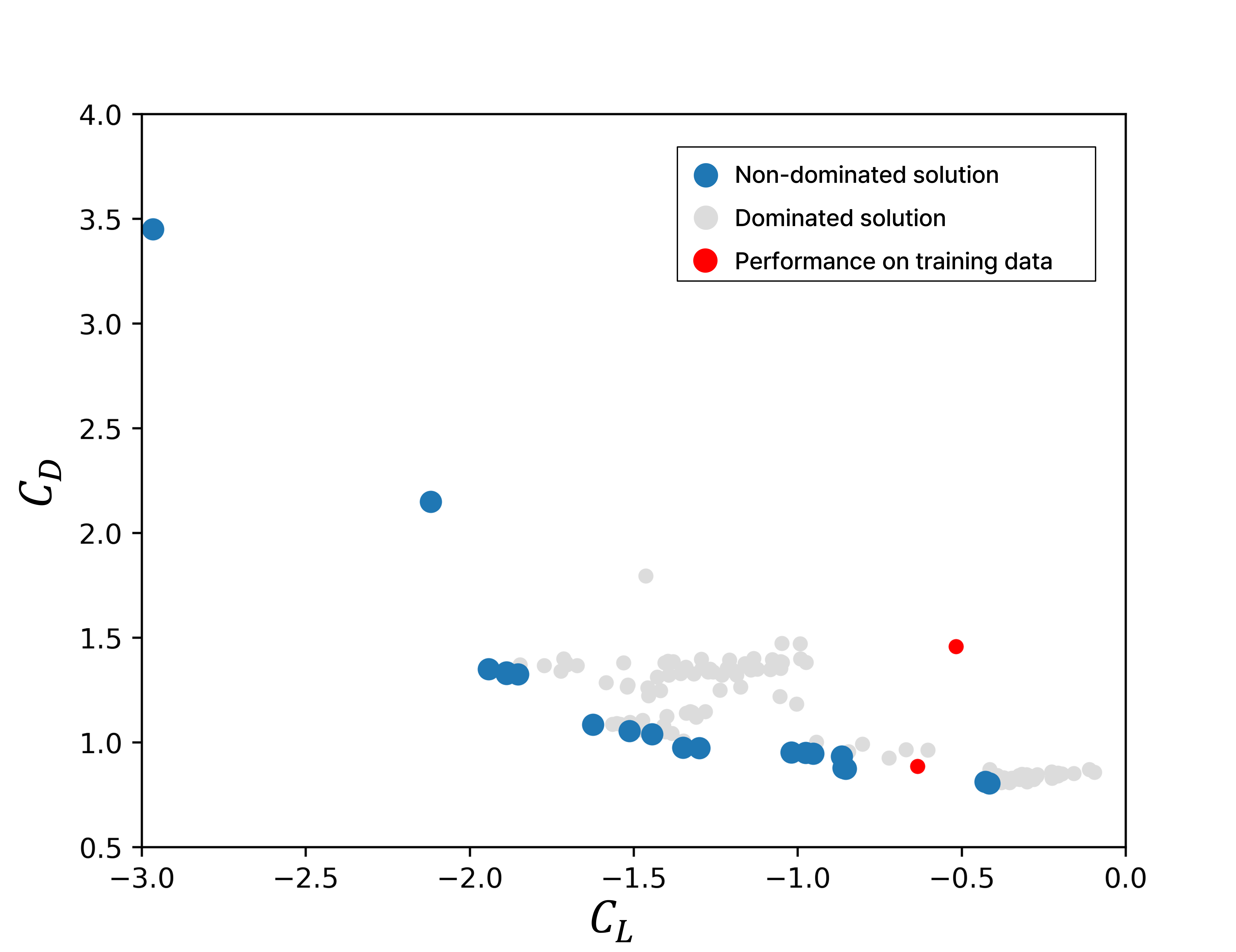}
                        \caption{{Pareto front results using two car samples}}
                        \label{fig:car2-pareto}
                    \end{subfigure}}
    \end{subfigure}
    \begin{subfigure}[tbh]{1.0\textwidth}
        \makebox[\textwidth][c]{
                    \begin{subfigure}[tbh]{0.5\textwidth}
                        \includegraphics[height=1.8cm]{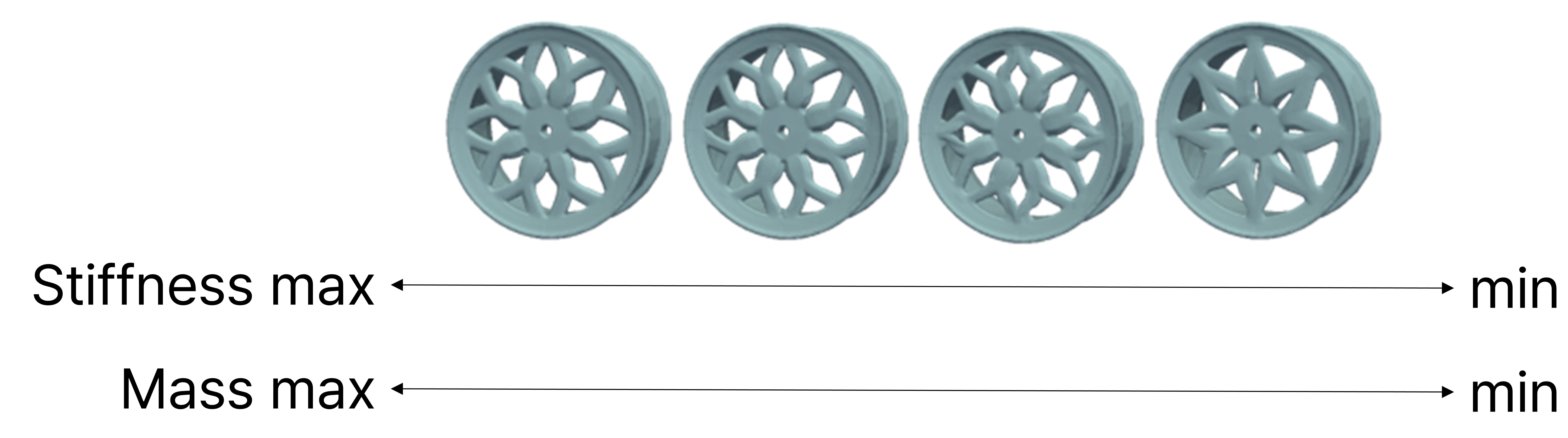}
                        \caption{{Non-dominated solutions from two wheel samples}}
                        \label{fig:wheel2-pareto-samples}
                    \end{subfigure}
                    \begin{subfigure}[tbh]{0.5\textwidth}
                        \includegraphics[height=1.8cm]{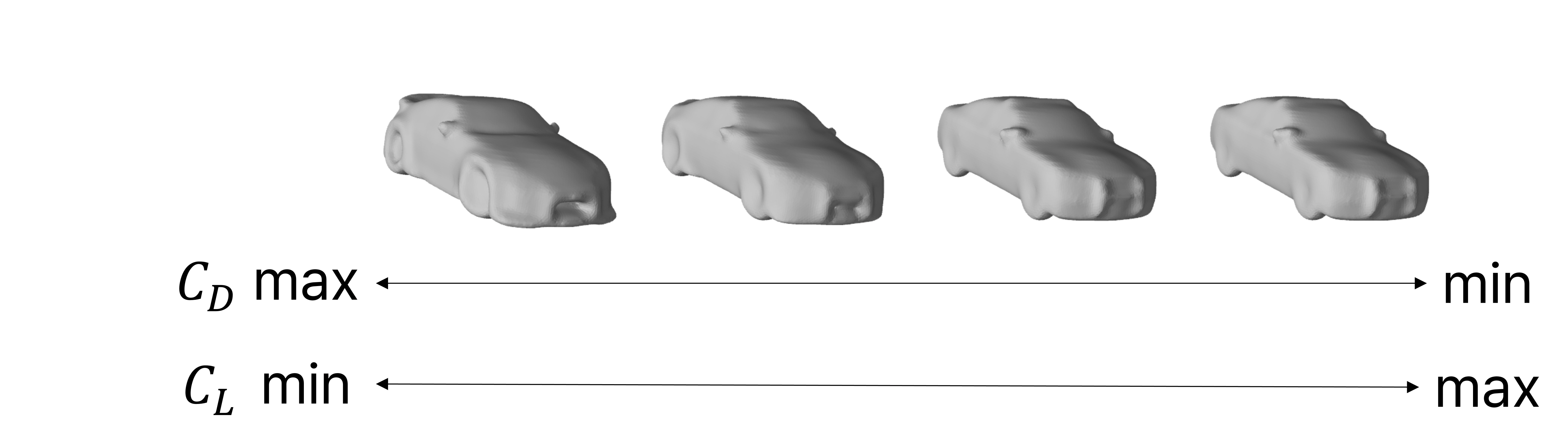}
                        \caption{{Non-dominated solutions from two car samples}}
                        \label{fig:car2-pareto-samples}
                    \end{subfigure}}
    \end{subfigure}
    \caption{Experimental results on {extremely limited data}}
\end{figure}

As shown in Figs.~\ref{fig:wheel2-pareto} and~\ref{fig:car2-pareto}, the optimization algorithm established a coherent Pareto front, even relying on just two data points for training. For the wheel shapes, the Pareto front based on stiffness and mass axes reveals that the model could generate diverse and engineering-relevant variations despite limited data. In Fig.~\ref{fig:wheel2-pareto-samples}, the model, trained on only two wheel designs, exhibits a smooth interpolation along the Pareto front, effectively capturing design variations as if performing parametric adjustments in stiffness and mass. This behavior indicates that the AI could simulate a continuum of structural adjustments between the limited initial designs, reflecting adaptability in exploring optimal shapes within minimal data constraints. Similarly, in the case of car shapes, the optimization process generated significant aerodynamic variations between the two baseline designs. As {shown} in Fig.~\ref{fig:car2-pareto-samples}, {the model maps meaningful shape changes that correlate with aerodynamic performance.} The algorithm effectively balanced the aerodynamic objectives and physical design limits, achieving a range of optimized forms that reflect varying levels of aerodynamic efficiency within the specified constraints.

These findings underscore the strength of {the proposed} methodology in handling datasets with limited diversity while still producing valuable and relevant design outputs. Although the initial data diversity is constrained, the model effectively explored the potential design space, extracting patterns and structural characteristics that offer insight into performance optimization. This investigation thus affirms the adaptability of computational design methods under conditions of sparse data, showcasing that, even with minimal input, significant design insights can be obtained.

\pagebreak

\section{{Conclusion}}
\label{sec:conclusion}

{This study proposes an engineering performance-based 3D shape optimization process that overcomes the drawbacks of existing methodologies while incorporating the advantages of conventional DGM-based shape optimization. Extensive hyper-parameter tuning across various architectures was conducted to select a suitable model for the proposed approach. By leveraging a simple regularization term and advanced encoding techniques, a model was developed that is well-suited for the limited dataset, a scenario typically considered one of the most challenging in the engineering field. Furthermore, three experiments were performed to highlight the advantages of this approach over conventional methods. The key findings of the study can be summarized as follows:}

\begin{figure}[h!]
    \makebox[\textwidth][c]{
        \begin{subfigure}[tbh]{0.6\textwidth}
            \centering
            \includegraphics[height=5.0cm]{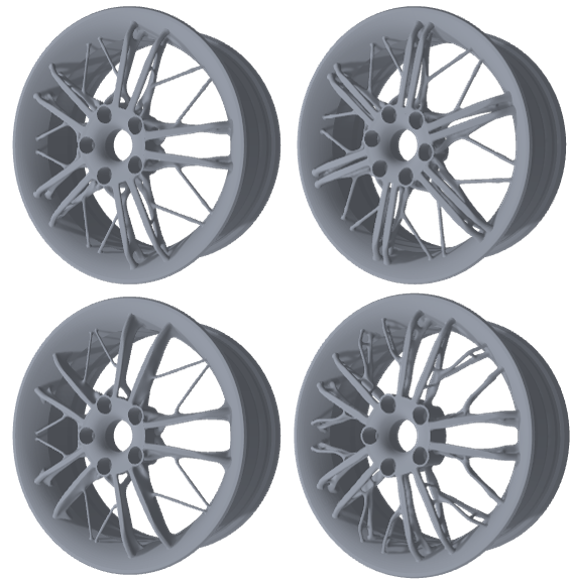}
            \caption{Generative design with topology optimization}
            \label{fig:to-wheel}
        \end{subfigure}
                
        \begin{subfigure}[tbh]{0.6\textwidth}
            \centering
            \includegraphics[height=5.0cm]{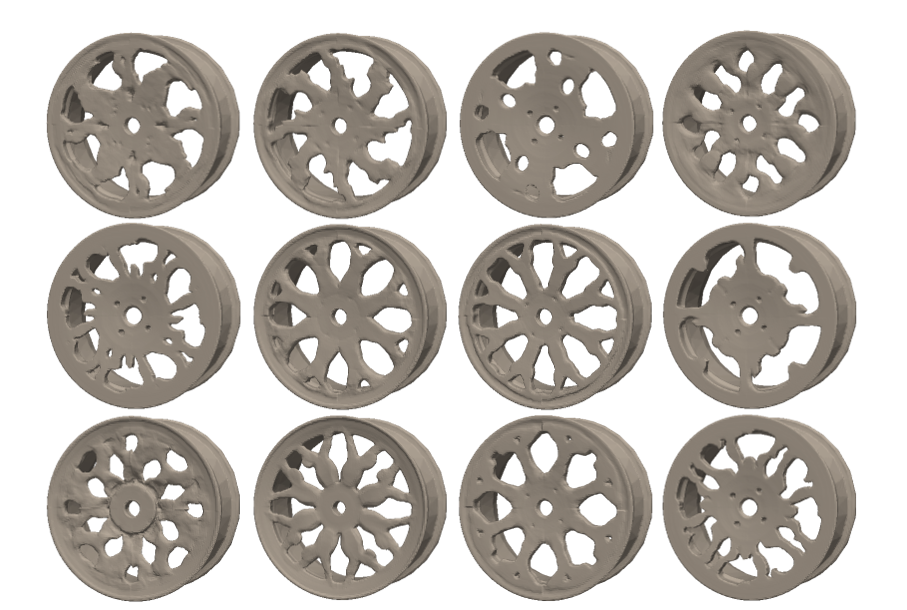}
            \caption{Generated shapes using the proposed method}
            \label{fig:dgm-wheel}
        \end{subfigure}}
    \caption{Comparison of shapes between the two methodologies}
    \label{fig:to-gd-comparison}
\end{figure}

\begin{enumerate}[label=\arabic*)]
    \item {Data Synthesis: Section~\ref{sec:wheel-exp30} details an experiment in which two distinct datasets are learned and optimized within the same latent space. This data-driven method, which does not require the explicit definition of parameters, enables exploration into previously undefined regions, where datasets Fig.~\ref{fig:to-data} and~\ref{fig:pd-data} are synthesized, illustrated in Fig.~\ref{fig:dgm-wheel}.}

    \item {Generalizability: Example experiments were conducted using wheel data for structural analysis and car data for CFD, two of the most commonly employed numerical analysis in shape optimization. Section~\ref{sec:exp} demonstrates that the optimization process successfully explores non-dominated solutions that outperform the performance of the actual data.}

    \item {Realizable Design: The proposed framework defines shapes by learning and extracting features from real data. As illustrated in Fig.~\ref{fig:to-wheel}, the data generated by topology optimization typically consist of a large amount of unstructured data with pronounced straight branches. In contrast, Fig.~\ref{fig:dgm-wheel} shows shape expressions that closely resemble those observed in real-world applications.}
\end{enumerate}

{The presented framework holds significant promise across two engineering applications. In domains such as car and wheel design, as employed in this study, the superiority of a shape cannot be defined solely in terms of engineering performance. These areas are inherently influenced by end-user preferences, with design iterations occurring through close collaboration between industrial designers and engineering designers who determine the overall form \unskip~\cite{jeon2024weaving}.}
{In this context, the proposed framework offers the advantage of exploring a data space that closely resembles real-world samples. This allows it to incorporate aesthetic considerations and manufacturability alongside quantitative engineering metrics. This capability paves the way for extending the framework to collaborative efforts within industrial design.}
{Moreover, the proposed approach can be integrated with recent advances in uncertainty quantification to develop task-aware surrogate models within an active learning paradigm. By defining objective values based on uncertainty rather than solely on engineering performance, the framework can facilitate more effective data sampling and contribute to the continuous improvement of surrogate model performance.}

{Despite these promising results, the proposed framework exhibits a notable limitation stemming from its strong dependence on training data. With ample data, the model can effectively generate a wide variety of design candidates. However, in scenarios where available training data are significantly limited—such as the extreme case discussed in Section~\ref{sec:exp2}—the performance range of the generated solutions becomes substantially constrained. Comparing performance distributions from Sections~\ref{sec:wheel-exp30}, \ref{sec:car-exp30} and \ref{sec:exp2}, it is evident that the distribution in Section~\ref{sec:exp2} covers only a small design region. While increased data allows the exploration of broader design spaces, limited data imposes inherent interpolation constraints, highlighting the critical importance of selecting appropriate training samples.}

{Moreover, the metrics used to evaluate generative models have limitations. Most prior studies assume that a sufficient volume of training data is available when defining their problems. Consequently, metrics such as CD, MMD and COV are well-suited for evaluating large datasets but may not be appropriate in the ``small data" regime defined here.
Due to the use of a quantitative shape evaluation, termed ``engineering performance”, the broad spread of performance distributions permits the inference that the latent space is smoothly structured. However, CD, MMD and COV values alone do not guarantee that a generative model has learned a meaningful latent representation. Therefore, there is a need for metrics capable of evaluating whether a meaningful latent space can be defined even when only a small amount of data is available.}

{In this context, augmenting the limited dataset with strategically meaningful data points is essential to enhance both generative and predictive model performance. Defining what constitutes ``meaningful data" for the model and empirically validating these definitions will be discussed as part of future research.}

\section*{CRediT authorship contribution statement}
\textbf{Yongmin Kwon}: Conceptualization; Data curation; Formal analysis; Investigation; Methodology; Resources; Software; Validation; Visualization; and Roles/Writing - original draft. \textbf{Namwoo Kang}: Project administration, Supervision, Investigation, Funding acquisition, Writing – Review \& Editing.

\section*{Acknowledgements}
This work was supported by the Ministry of Science and ICT of Korea grant (No. 2022-0-00969, No. 2022-0-00986, and No. GTL24031-000) and the Ministry of Trade, Industry \& Energy grant (RS-2024-00410810).

\bibliographystyle{elsarticle-num} 
\bibliography{cas-refs}






\end{document}